\pdfoutput=1
\documentclass[11pt]{article}

\usepackage[preprint]{acl}

\usepackage{times}
\usepackage{latexsym}
\usepackage{enumitem}
\usepackage{pifont}
\usepackage{balance}

\usepackage[T1]{fontenc}
\usepackage[utf8]{inputenc}

\usepackage{microtype}

\usepackage{inconsolata}

\usepackage{graphicx}
\usepackage{amsmath}
\usepackage{booktabs}
\usepackage{algorithm}
\usepackage{algorithmic}
\usepackage{caption}
\usepackage{subcaption}
\usepackage{multirow}
\usepackage{multicol}
\usepackage{xcolor}
\usepackage{colortbl}
\usepackage{adjustbox}
\usepackage{textcomp}
\usepackage{cleveref}

\newcommand{\sld}[1]{%
{\textcolor{black}{#1}}%
}
\newcommand{\added}[1]{%
{\textcolor{black}{#1}}%
}
\expandafter\def\expandafter\normalsize\expandafter{%
\normalsize%
\setlength\abovedisplayskip{4pt}%
\setlength\belowdisplayskip{4pt}%
\setlength\abovedisplayshortskip{-8pt}%
\setlength\belowdisplayshortskip{0pt}%
}
\newcommand{\model}{\textsc{KcMF}}

\newcommand{\SMAT}{\textsc{smat}}
\newcommand{\Jellyfish}{\textsc{JellyFish}}

\colorlet{newtab}{white}
\colorlet{newtabemph}{blue!50}
\colorlet{newtabcap}{black}

\title{\model{}: A Knowledge-compliant Framework for Schema and Entity Matching with Fine-tuning-free LLMs}

\author{
 \textbf{Yongqin Xu\textsuperscript{1}},
 \textbf{Huan Li\textsuperscript{1,2}},
 \textbf{Ke Chen\textsuperscript{1,2}},
 \textbf{Lidan Shou\textsuperscript{1,2}}
\\
 \textsuperscript{1}State Key Laboratory of Blockchain and Data Security, Zhejiang University \\
 \textsuperscript{2}Hangzhou High-Tech Zone (Binjiang) Institute of Blockchain and Data Security
\\
\texttt{\{xuyongqin\}@zju.edu.cn}
}

\begin{document}
\maketitle
\begin{abstract}
Schema matching~(SM) and entity matching~(EM) tasks are crucial for data integration.
While large language models (LLMs) have shown promising results in these tasks, they suffer from hallucinations and confusion about task instructions. 
This study presents the Knowledge-Compliant Matching Framework (\model{}), an LLM-based approach that addresses these issues without the need for domain-specific fine-tuning.
% \model{} employs a pseudo-code-based task decomposition strategy to generate task-specific conditional statements that guide LLM reasoning and reduce confusion. 
\model{} employs a \added{once-and-for-all} pseudo-code-based task decomposition strategy to adopt natural language statements that guide LLM reasoning and reduce confusion \added{across various task types.} 
We also propose two mechanisms, Dataset as Knowledge~(DaK) and Example as Knowledge~(EaK), to build domain knowledge sets when unstructured domain knowledge is lacking. Moreover, we introduce a result-ensemble strategy to leverage multiple knowledge sources and suppress badly formatted outputs.
% a pseudo-code-guided matching framework with knowledge enhancement, the Knowledge-compliant Matching Framework (\model{}), to solve matching tasks with fine-tuning-free LLMs. To tackle the above issues, we introduce a pseudo-code based strategy to disambiguate the task definition, and we propose the Dataset as Knowledge~(DaK) and Example as Knowledge~(EaK) mechanisms to construct domain knowledge sets for task domains in scarcity of natural language formed knowledge. 
\sld{Extensive evaluations confirm that \model{} clearly enhances five LLM backbones in both SM and EM tasks while outperforming the non-LLM competitors by an average F1-score of 17.93\%.}
% showing \model{} outperforms previous non-LLM SoTA by an average F1-score of 22.9\%, and can compete with the SFT-LLM SoTA.
\end{abstract}

\section{Introduction}

Schema matching is the task of identifying correspondences between elements of two or more database schemas. This task plays an important role in data integration efforts. Another task called entity matching, also known as entity resolution or record linkage, aims to identify schema instances that refer to the same real-world entity. While these two tasks vary in their definitions and approaches, they share the common goal of matching database elements. This paper attempts to tackle these tasks under the umbrella term \emph{data matching}, employing a unified methodology while still retaining their original problem settings independently, 
%In this study, we term schema matching and entity matching collectively as data matching, motivated by the same goal of matching database elements shared by these two tasks. The exploration in this paper brings the solutions of two tasks under a unified methodology but still retains their settings independently, 
i.e., entity matching is solved based on the database records, whereas schema matching is solved using metadata only due to privacy considerations.
	
 \begin{figure}[t]
\centerline{\includegraphics[width=\columnwidth]{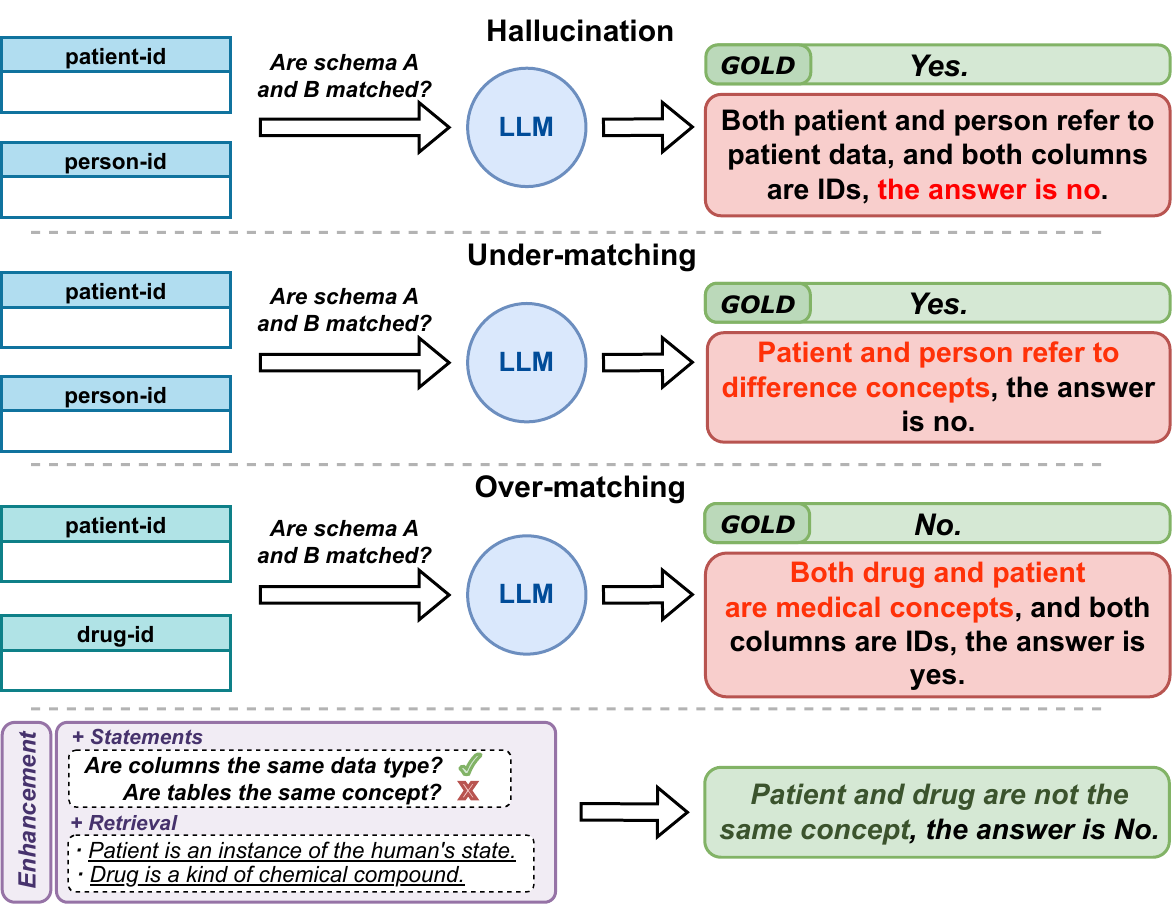}}
\caption{Three common issues in LLM-based data matching tasks and an overview of the enhancement will be discussed in this study: As demonstrated at the bottom, with pseudo-code and retrieved knowledge, by going through statements from the former, the LLM are able to reject the match between \emph{patient-id} and \emph{drug-id}.
% (1) \emph{Hallucination}: The LLM generates incorrect deductions even when provided with correct evidence.
% (2) \emph{Under-matching}: The LLM rejects valid matches with overly strict criteria.
% (3) \emph{Over-matching}: The LLM over-generalizes and incorrectly associates input data. 
% The enhancement proposed in this paper is depicted at the bottom of this figure.
}
\label{figure:matching_errors}
\end{figure}
 
Early systems for data matching tasks rely on expert systems or traditional machine learning methods, while modern data matching approaches often employ pre-trained language models (PLMs) as the backbone~\citep{li_deep_2020,bellatreche_smat_2021,zeakis_pre-trained_2023}, which is then fine-tuned on task-specific datasets. 
% due to their strong text comprehension capabilities. 
However, PLM-based methods \sld{usually incur efforts to collect} fine-tuning data and face performance degradation when handling out-of-domain data during inference. 
Large Language Models (LLM) 
%or Foundation Models (FM) 
are PLMs of massive scales with billions of parameters, trained on vast and diverse datasets. 
With huge internal knowledge gained from pre-training and strong representation capability, LLMs show competitive performance on numerous data-wrangling benchmarks, particularly without the need for fine-tuning~\citep{narayan2022can,peeters2023using,sheetrit_rematch_2024,peeters_entity_2024}.
This remarkable efficacy liberates users from the burden of fine-tuning efforts.

Despite promising results, LLM-based methods for data matching face several performance issues. 
For clarity of presentation, we focus mainly on schema matching in the main text; further discussions of entity matching are provided in Appendix~\ref{appendix-section:methodology}.
% Although LLM-based methods achieve promising results on matching tasks, their performance is restricted due to a few problems observed in either LLM or the matching tasks running atop. 
As depicted in Figure~\ref{figure:matching_errors}, these issues are categorized into three types: (1) \emph{hallucination}, where the LLM generates incorrect deduction from correct evidence at hand; (2) \emph{under-matching}, where the LLM rejects matching with overly strict criteria; and (3) \emph{over-matching}, where the LLM over-generalizes the association of the input data. 
While hallucination is a widely recognized problem that can harm the LLM's performance in many other tasks~\citep{zhang_sirens_2023,rawte_survey_2023}, under-matching and over-matching, jointly termed as \textbf{\emph{confusion}} in this paper, are challenges specific to the data matching tasks.
%, and always lead to failure in matching tasks.

This study presents the Knowledge-compliant Matching Framework (\model{}), a fine-tuning-free and retrieval-enhanced approach to data matching tasks. 
To address the problem of confusion, \model{} employs a \emph{pseudo-code-based task decomposition strategy} for LLMs. 
Pseudo-code here is an ordered list of task-specific conditional statements, written in natural language, that guide the matching process.
The LLM can easily evaluate the validity of each condition for the data being matched, and then follow the corresponding directives.
% for the matching task. The validity of each condition in these statements can be easily evaluated by the LLM, and their respective directives are then taken. 
By walking through the pseudo-code, the LLM is able to reason its way to the final matching result. 
\sld{Unlike Chain-of-Thought~\citep{wei2022chain}, which relies on LLM's internal knowledge to generate reasoning steps, \model{} uses explicit task-aware pseudo-code, eliminating the need to engineer prompts for different datasets}\added{, that is, pseudo-code is written only once for a task~(such as schema matching).}
% takes this process under the guidance of task-specific pseudo-code, eliminating the need for writing thoughts across datasets. 

To address hallucination, \model{} incorporates external knowledge in the form of natural language sentences related to the input data. 
To compensate for the lack of such unstructured domain knowledge, \model{} builds domain knowledge sets by leveraging 
readily available datasets (\emph{Dataset as Knowledge}, DaK) and examples (\emph{Examples as Knowledge}, EaK) from various domains. 
Since fine-tuning-free LLMs tend to generate improperly formatted outputs that do not match the format given in the demonstration, we employ a technique called \emph{Inconsistency-tolerant Generation Ensemble} (IntGE) to suppress such unexpected outputs and maintain an automated downstream workflow. 

\sld{We evaluate \model{} on three SM datasets and four EM datasets. The comprehensive results show that \model{} significantly improves five LLM backbones, exceeds the current non-LLM best-performers (\SMAT{}~\citep{bellatreche_smat_2021} for SM and \textsc{sudowoodo}~\citep{wang_sudowoodo_2023} for EM), and achieves performance comparable to fine-tuned LLMs in certain cases, all without any fine-tuning.}

% \sld{We evaluate \model{} using 3 knowledge-intensive datasets from OMAP project~\citep{bellatreche_smat_2021} for SM, and \textsf{MMM}, modified from MedMentions~\citep{mohan_medmentions_2019} together with 3 commonly-used datasets from DeepM~\citep{mudgal_deep_2018} for EM. The results on these 7 datasets show that \model{} surpasses 1) all LLM baselines, 2) the respective non-LLM SoTAs --- SMAT{}~\citep{bellatreche_smat_2021} and \textsc{sudowoodo}~\citep{wang_sudowoodo_2023} without any fine-tuning efforts. Our main contributions are as follows.}
\begin{enumerate}[leftmargin=*,itemsep=-0.5em,topsep=0em]
\item We introduce \model{}, an LLM-based framework that incorporates a unified methodology to solve schema matching and entity matching tasks, eliminating the need for task-dependent model design and domain-specific fine-tuning.
% designed to solve data matching tasks without the need for domain-specific fine-tuning, putting data matching efforts under a unified procedure.
% domain-specific fine-tuning and done a comprehensive evaluation showing \model{} outperforms all LLM baselines and the non-LLM SoTA \SMAT{}~\citep{bellatreche_smat_2021}.

\item We propose 1) \sld{a once-and-for-all pseudo-code-based task decomposition strategy across task types, to guide the LLM's reasoning by obtaining task-specific conditional statements, reducing \emph{confusion} during matching}, 2) two mechanisms, namely DaK and EaK, to build knowledge sets in lack of unstructured domain knowledge, and 3) IntGE, a result-ensemble strategy designed to leverage multiple knowledge sources and suppress poorly formatted outputs.

\item We perform a comprehensive evaluation for \model{} and demonstrate its superiority against various popular SM and EM methods. Our analysis also shows that \model{} can generalize effectively across different LLMs.
\end{enumerate}

\section{Background and Task Definition}

\subsection{Related Work}

Conventional \textbf{\emph{Schema Matching}} (SM) approaches can generally be categorized into 1) \emph{constraint-based methods}, which utilize attributes defined in database constraints to measure similarity among schemas~\citep{alexe2010mapmerge,chen_biggorilla_2018,atzeni_meta-mappings_2019}, and 2) \emph{linguistic-based methods}, which leverage the semantic information contained in schema names or descriptions to construct mappings~\citep{kettouch_using_2017,asif-ur-rahman_semi-automated_2023}. 
More recent studies have adopted deep neural networks to tackle SM. 
\citet{bellatreche_smat_2021} use an attention-based BiLSTM with pre-trained word embeddings. 
\citet{zhang_schema_2023} leverage BERT~\citep{kenton2019bert} to generate schema features, which are then used to train a linear classifier under semi-supervised learning. 
With the advent of LLMs, \citet{narayan2022can} were the first to apply LLMs to SM, using straightforward serialization with few-shot settings. 
Recent LLM-based SM approaches, such as \citet{zhang_jellyfish_2024}, attempt to tackle this task using Supervised Fine-tuning (SFT).
While SFT achieves strong results, it needs tremendous efforts to collect fine-tuning data and is hard to transfer to unseen domains. 
Another line of works, such as~\citet{sheetrit_rematch_2024} and~\citet{parciak_schema_2024}, utilize proprietary LLMs, which have shown promising results on several benchmarks. However, these approaches still suffer from high computational costs and produce indecisive outputs. 
Our approach aims to extend the scope of LLM-based SM by utilizing task-specific pseudo-code to guide LLMs' predictions within a single-round inference. 
The design offers explicit criteria for the matching task, addressing the {confusion} problem mentioned earlier.

\begin{figure*}[ht]
\centerline{\includegraphics[width=\textwidth]{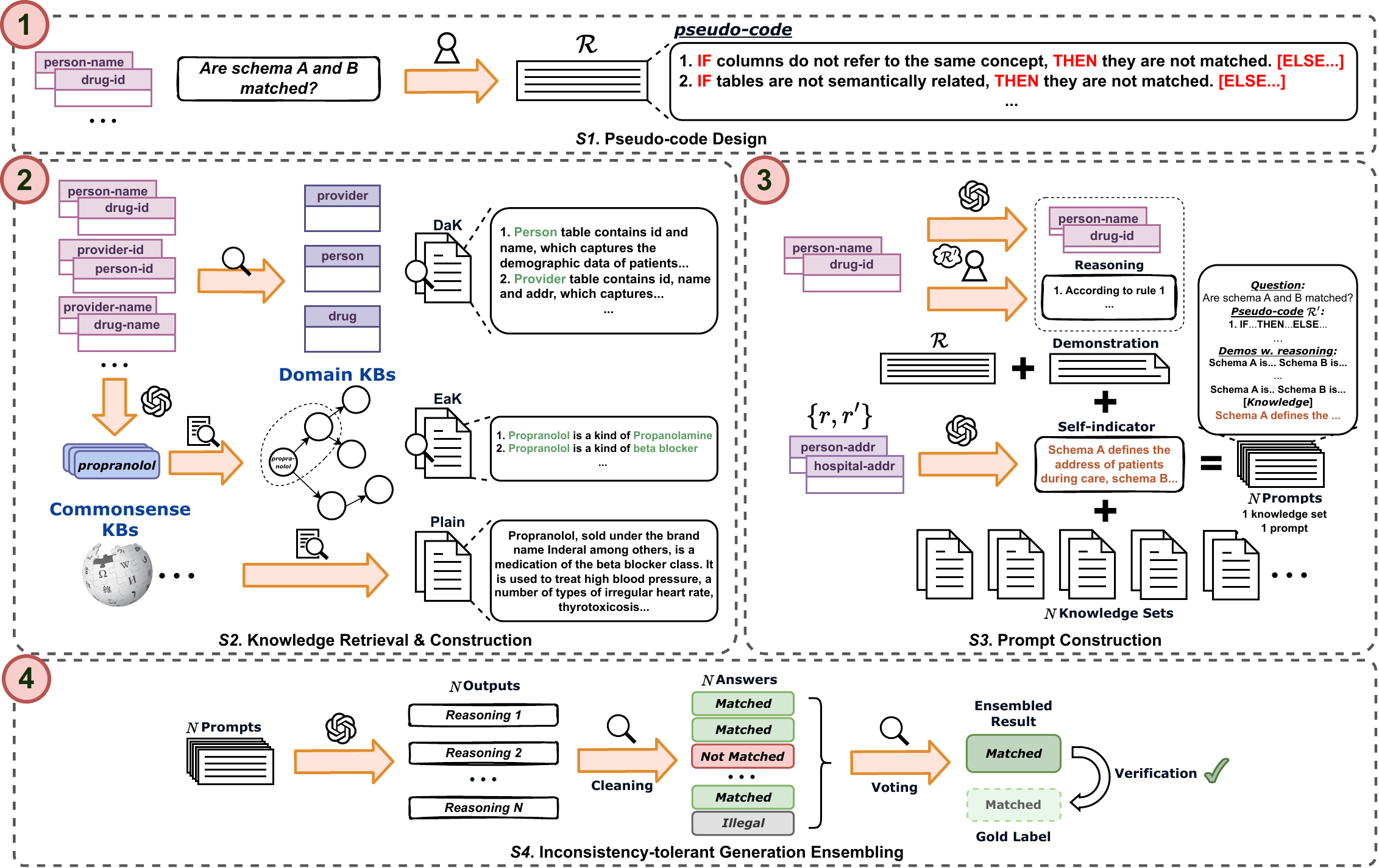}}
\caption{Overview of \model{}. Our carefully designed pseudo-code (detailed in Appendix~\ref{appendix-section:implementation}) offers a reusable and efficient solution for both SM and EM tasks. \sld{This eliminates the need for redesigning statements from scratch when working with new datasets, streamlining the process and enhancing adaptability.}}
\label{figure:main}
\end{figure*}

\smallskip
\noindent\textbf{\emph{Entity Matching}} (EM) is often considered a downstream task of SM by many previous works \citep{nie_deep_2019,li_deep_2020,brunner2020entity,barlaug2021neural}. 
Similar to SM, traditional non-LLM solutions for EM focus on computing similarity between entities~\citep{thirumuruganathan_reuse_2018,ebraheem_deeper_2018,kasai2019low,li_deep_2020}.
Following the initial application of LLM to EM~\citep{narayan2022can}, LLM-based solutions have been actively explored~\citep{fan2024cost,peeters_entity_2024,li2024booster,wang_match_2024}. 
From the outset, our approach is designed as a unified framework capable of addressing both SM and EM.
By recognizing the deep semantic similarities and shared motivations between SM and EM --- such as aligning and matching data elements --- we develop a methodology that seamlessly integrates both tasks. 
% The approach in this paper was originally formed on SM, but due to the semantic similarities between the motivations of SM and EM, and the possibility of the two tasks being solved under the same methodology, this paper treats EM as a target task on par with SM.

\smallskip
\noindent\textbf{\emph{Retrieval Augmented Generation}} (RAG) incorporates retrieved documents into queries and has become a popular paradigm in mitigating hallucination in LLMs~\citep{lewis2020retrieval,yu_chain--note_2023,shao_enhancing_2023,jiang_active_2023,asai_self-rag_2023,xu_recomp_2023,shi2024replug}. 
While RAG has shown reliable results in addressing hallucination, its performance is limited by the quality of the retrieved documents and, obviously, is challenging to deploy in scenarios lacking unstructured knowledge sources.
Our approach adopts the concept of introducing external knowledge to alleviate hallucination. 
In particular, to address the lack of unstructured domain knowledge in data matching,
we present two mechanisms, DaK and EaK, for building knowledge sets by utilizing existing domain knowledge bases. 
Our approach showcases the potential of external knowledge enhancement strategies on LLM-based classification tasks.

\subsection{Task Definition}
\label{section:problem_statement}

We denote by $\mathcal{S}$ the \emph{source schema} from database $\mathcal{D}$, and $\mathcal{S'}$ the \emph{target schema} from database $\mathcal{D}'$.
The goal of \textbf{schema matching} (SM) is to identify all pairs of \textbf{attributes} $(\mathcal{A}, \mathcal{A}')$ such that  $\mathcal{A} \in \mathcal{S}$ and $\mathcal{A}' \in \mathcal{S}'$, and both attributes
represent the same information in their respective schemas. 

In this paper, we focus on a more straightforward scenario: we enumerate all possible mappings $\mathcal{M} \subseteq \mathcal{D} \times \mathcal{D'}$ and determine whether each mapping $\{r, r'\} \in \mathcal{M}$ is correct or not. 
Each candidate mapping $\{r, r'\}$ consists of two items, where each item $r$ is composed of a schema name $N$ and a schema description $C$ from the corresponding database $\mathcal{D}$.

We framework our schema matching task as an LLM generation task guided by pseudo-code instructions under a knowledge-enhanced setting.
This involves 1) a list $\mathcal{P}$ of designated task-specific pseudo-code and 2) a list $K$ of knowledge items retrieved from all available knowledge sets $\mathcal{K}$.

% Specifically, in this paper, given a candidate mapping $\{r,r'\}$, denoted as
% \begin{equation}			
% \label{equation:mappings}
% \{r, r'\} = \{(N, C), (N', C')\},
% \end{equation}
% where each of the two items consists of a schema name $N$ and a schema description $C$. 
% And together with $\mathcal{P}$ and $K$, $\{r,r'\}$ are fed into LLM $\mathcal{L}$ to generate a response $\mathcal{L}_{opt}$:

We serialize the inputs, including the pseudo-code $\mathcal{P}$, candidate mapping $\{r,r'\}$, and retrieved knowledge items $K$, into a prompt tailored to the LLM $\mathcal{L}$ using a function $\phi$.
An LLM-generated response $LR$ is then obtained as: 
\begin{equation}
\label{equation:definition}
LR \gets \mathcal{L}\big(\phi(\mathcal{P}, \{r,r'\}, K)\big).  
\end{equation}
Our task objective is to classify the correctness of each candidate mapping. Specifically, we aim to obtain a binary classification result $c \in \{yes, no\}$ from the LLM's response $LR$, indicating whether the mapping $\{r,r'\}$ is correct. For clarity of presentation, here we focus on defining SM, though it should be noticed that this setting can also be generalized to entity matching task (cf.\ Appendix~\ref{appendix-section:methodology}).

% where $\phi$ is a function for serialization. Thus, the task objective is to classify the correctness of each candidate mapping. Formally, to get a binary classification result from the response of an LLM: $c \leftarrow \mathcal{L}_{opt}$, where $c \in \{yes, no\}$. 

% Similarly, for entity matching, the goal is to determine if the entities mapping $\{e, e'\} \in \mathcal{D} \times \mathcal{D'}$ is correct, or to judge whether entities $e$ and $e'$ refer to the same real-world entity or not.

% To form a unified notation for the matching task, we define a new variable representing the input mappings, combining schema matching and entity matching tasks:

% \begin{equation}			
% \{r, r'\} = \begin{cases}
% \{e, e'\} \in \mathcal{D} \times \mathcal{D'} \\
% \{S, S'\} \in \mathcal{D} \times \mathcal{D'}
% \end{cases}
% \end{equation}

% Thus, given a mapping $\{r, r'\}$ between $\mathcal{D}$ and $\mathcal{D'}$, the objective of the matching task is to classify the correctness of the given mapping.

\section{Matching Framework}
\label{section:matching_framework}

\subsection{Framework Overview}

Figure~\ref{figure:main} presents an overview of our proposed \model{}, which operates in four sequential stages.
\emph{\textbf{S1}. Pseudo-code Design}: Experts decompose the task into task-related conditional statements in natural language. A superior advantage of our approach is that this pseudo-code for a task is designed once and can be reused by anyone performing the matching task, without the need to write custom ones.
\emph{\textbf{S2}. Knowledge Retrieval \& Construction}: 
\model{} constructs granular domain knowledge and database-structure knowledge via retrieving information from domain knowledge bases and discovering database structures. 
\emph{\textbf{S3}. Prompt Generation}:
Utilizing the knowledge from S2 and pseudo-code from S1, \model{} generates LLM prompts that include demonstrations.
% which uses knowledge retrieved from Stage (2) and pseudo-code designed in Stage (1) to generate prompts with demonstrations. 
\emph{\textbf{S4}. Inconsistency-tolerant Generation Ensemble}: \model{} uses the constructed prompts to query the LLM.
The multiple outputs generated are combined, mitigating the ill-formatting issue, to reach the final decision. 

Illustrative examples of the main \model{} components in this section are provided in Appendix~\ref{appendix-section:methodology}.

\subsection{Pseudo-code Design}   
\label{section:decomposing_complicated_task_into_rules}
	
As Figure~\ref{figure:matching_errors} depicts, one of the challenges that LLMs face when performing matching is the issue of under-matching and over-matching.
This issue stems from the ambiguity of the task instruction ``\texttt{match}''.
Without additional context, the word \texttt{match} lacks a clear definition in this setting\footnote{As shown in Appendix~\ref{appendix-section:gpt_response_sm}, there is a discrepancy between what \texttt{GPT-3.5} understands the instruction \texttt{match} and the intended objective of the schema matching task.}. 
Take schema matching as an example.
The term \texttt{match} in this context has at least three different interpretations, depending on the focus of the task:
\begin{itemize}[leftmargin=*,itemsep=-0.5em,topsep=0em]
	\item The \emph{data types} of the column values are \emph{the same} or \emph{convertible} to each other.
    \item Table definitions are \emph{semantically related}.
    \item Schemas refer to \emph{the same real-world concept}.
\end{itemize}
\sld{To address this ambiguity, we propose a task-aware strategy to decompose the task into pseudo-code composed of conditional statements directly derived from the task's motivation.}
As shown in S1 of Figure~\ref{figure:main}, natural language predicates are structured into \emph{if-then-else} constructs, each providing sufficient conditions to determine ``matched'' or ``not matched'' cases.
The pseudo-code is designed to comprehensively cover all relevant task conditions. 
This process requires manual effort to understand the task's motivation and design the pseudo-code~$\mathcal{R}$ (see implementation details in Appendix~\ref{appendix-section:implementation}).
To apply this pseudo-code, we introduce a reasoning prompting strategy inspired by Chain-of-Thought~\citep{wei2022chain}, which will be discussed in~\cref{section:prompt_construction}.
	
\subsection{Building Domain Knowledge Set}
\label{section:domain_knowledge_set}
	
To solve the matching task within a knowledge-enhanced setting, we first need to retrieve a domain knowledge list $K$. 
% For common sense QA tasks, there are existing large-scale natural language corpora such as Wikipedia, but domain knowledge bases (KBs) are usually entity-based or in the form of a thesaurus, where natural language knowledge is hard to retrieve.
However, because answers for matching tasks cannot be explicitly derived from retrieved information and due to the scarcity of unstructured domain knowledge, we propose two mechanisms: Dataset as Knowledge (DaK) and Examples as Knowledge (EaK). These mechanisms are designed to construct highly relevant unstructured knowledge sets tailored specifically for matching tasks. Further implementation details are provided in Appendix~\ref{appendix-section:implementation}.
	
\begin{figure}[htbp]
\centerline{\includegraphics[width=\columnwidth]{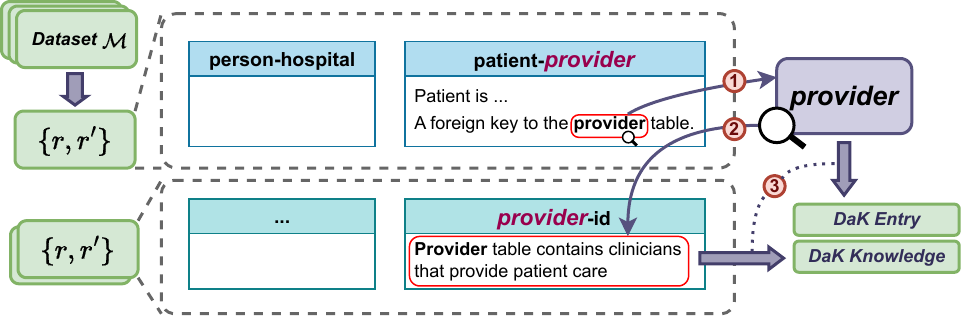}}
\caption{A toy example of DaK. Object ``provider'' and the description of which are identified, respectively (\ding{172} \& \ding{173}); then, they are integrated as an entry and a piece of DaK knowledge (\ding{174}).
% A toy example of DaK. The discovered object ``provider'' is reformed as an entry, and the metadata of ``provider'' discovered from another example is extracted as the knowledge.
}
\label{figure:dak}
\end{figure}

\paragraph{Dataset as Knowledge (DaK)}
Due to privacy concerns, concrete records from the source databases $\mathcal{D}$ and $\mathcal{D'}$ are often inaccessible to the schema matching system~\citep{johnson_mimic-iv_2023,zhang_schema_2023}.
This means that the matching task must be performed using only metadata. 
In this setting, LLMs are required to understand the structures of $\mathcal{D}$ and $\mathcal{D'}$, which, however, can only be partly seen during inference through the limited metadata representations $\{r, r'\}$ (cf.\ Section~\ref{section:problem_statement}). 
Hence, we propose DaK to acquire knowledge of the structure of source databases by searching metadata from the full dataset $\mathcal{M}$, as shown in Figure~\ref{figure:dak}. 
% It consists of two procedures as follows.

The procedure of DaK aims to discover database objects and their metadata from the candidate pool $\mathcal{M}$ (see pseudo-code in Appendix~\ref{appendix-section:methodology}). 
As shown in Figure~\ref{figure:dak}, DaK operates in three steps:
1) for a given candidate pair $\{r,r'\} \in \mathcal{M}$, DaK identifies an object $O$; in the example, $O$ is a table named ``provider''. While this step can utilize techniques like Named Entity Recognition (NER), for rapid prototyping, we extract $O$ by traversing all schema names $N$ and $N'$ associated with ${r, r'}$ (see \ding{172}).
2) DaK then scans the remaining pairs $\{r,r'\}$ to find metadata $K_{\text{DaK}}$ related to the identified object $O$; in this example, $K_{\text{DaK}}$ is the description of the table ``provider''. This is accomplished by matching $O$ with descriptions $C$ from the other candidate pairs (see \ding{173}). 3) After obtaining an object list $\mathcal{C}_{\text{Obj}}$ consisting of each identified object $O$, and a metadata list $\mathcal{C}_{\text{DaK}}$ consisting of all matched knowledge $K_{\text{DaK}}$, DaK proceeds to form the name of $O$ as an entry and $K_{\text{DaK}}$ of $O$ as the corresponding DaK knowledge. 
As shown in Figure~\ref{figure:dak}, the name of the identified object ``provider'' and the description of ``provider'' are constructed into a piece of DaK knowledge: \{\textit{provider}: \textit{provider table contains clinicians that provide patient care}\}.

\begin{figure}[htbp]
	\centerline{\includegraphics[width=\columnwidth]{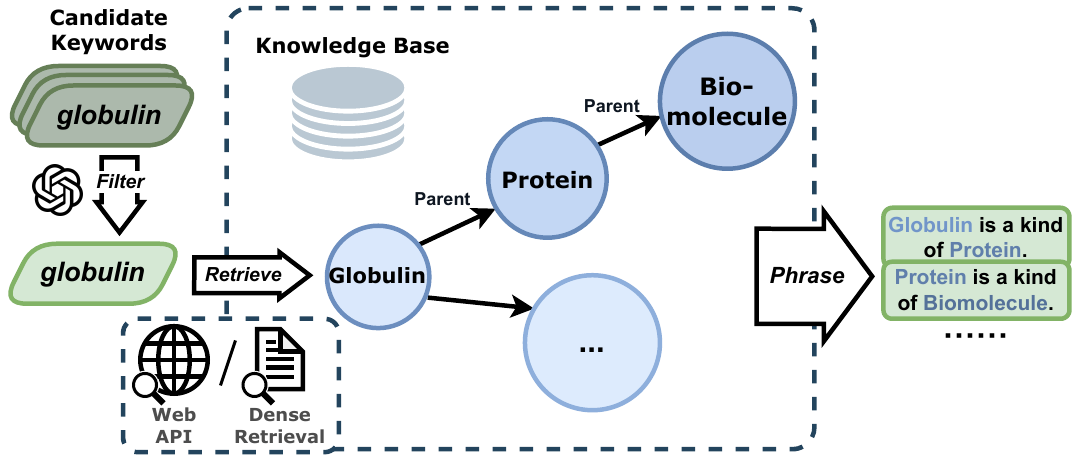}}
		\caption{An example of EaK.}
		\label{figure:eak}
\end{figure}

\paragraph{Examples as Knowledge (EaK)}
Using text chunks from a self-built corpus has become a \textit{de facto} approach for retrieval enhancement to reduce LLM hallucinations~\citep{gao_retrieval-augmented_2024}. 
For common-sense QA tasks, such text chunks can be easily retrieved from existing commonsense corpora. 
However, the knowledge required for matching tasks is domain-specific, and existing KBs are mainly structured as entity databases or thesaurus rather than natural language text. 
Thus, we propose EaK, as depicted in Figure~\ref{figure:eak}.
% To this scarcity, EaK is designed to construct a natural language domain knowledge set.
		
As its name suggests, EaK aims to explain complex concepts using examples. 
Given a pair $\{r,r'\}$, EaK first extracts keywords from it by querying an LLM, and then queries an LLM again to filter the domain-irrelevant ones (see pseudo-code in Appendix~\ref{appendix-section:prompts}). Obtaining all candidate keywords, EaK uses these candidates to search domain KBs for the top-$k$ related records. For remote KBs, retrieval can be done through a search over the provided Web API; while for local KBs, this procedure can be implemented using stronger dense retrieval. 
For each retrieved record, EaK leverages its relationships and properties to form explanatory knowledge. For example, in Figure~\ref{figure:eak}, EaK generates the knowledge by phrasing the parent relation.
% presented in Example~\ref{example_eak}.
% \begin{example}
% \label{example_eak}
% EaK knowledge examples:  \\
% \centerline{example: A \{Globulin\} is a kind of \{Protein\}. } \\
% \centerline{example: A \{Protein\} is a kind of \{Biomolecule\}.} 
% \end{example}
	
\subsection{Prompt Construction}
\label{section:prompt_construction}

\label{section:rule_based_reasoning_prmopting}
\begin{figure}[htbp]
    \centerline{\includegraphics[width=\columnwidth]{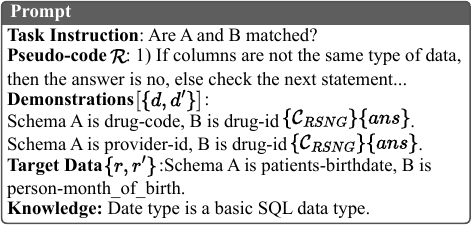}}
    \caption{A toy prompt combining all outcomes from previous sections and placeholders $\{\mathcal{C}_\text{RSNG}\}$ and $\{ans\}$.}
    \label{figure:toy_prompt}
\end{figure}

\paragraph{Pseudo-code-based Reasoning Prompting}
In a paradigm of $k$-shot in-context learning, we sample $k$ valid match pairs $\{d, d'\}$ as demonstrations together with the pseudo-code $\mathcal{R}$, the queried pair $\{r, r'\}$, and knowledge list $K$. 
The fundamental pattern of the prompt is illusrated in Figure~\ref{figure:toy_prompt}.

	% \begin{example}
	% 	\label{example:fundamental_prompt_pattern}
	% 	A prompt of the fundamental pattern:	
	% 	\begin{center}
	% 		\textbf{Task Instruction}							\\
	% 		Are A and B matched?			\\
	% 		\textbf{Pseudo-code}~$\mathcal{R}$							\\ 
	% 		If columns are not the same data type, then the answer is no, else check the next statement. \\
	% 		...\\
	% 		\textbf{Demonstrations}~$[\{d,d'\}]$						\\  
	% 		Schema A is provider-id, B is drug-id $\{\mathcal{C}_\text{RSNG}\}\{ans\}$\\
	% 		Schema A is drug-code, B is encounter-id $\{\mathcal{C}_\text{RSNG}\}\{ans\}$\\
	% 		\textbf{Target Data}~$\{r,r'\}$						\\ 
	% 		Schema A is patient-id, B is person-id...				\\
	% 		\textbf{Knowledge}							\\ 
	% 		ID is a database identifier \\
	% 	\end{center}
	% \end{example}
	% \iffalse
	% 	\rule{0.75\columnwidth}{0.5pt} \\
	% \fi
	
There are two placeholders in the prompt, $\{\mathcal{C}_\text{RSNG}\}$ for the reasoning steps and $\{ans\}$ for the answer. 
To generate $\mathcal{C}_\text{RSNG}$ and $ans$, we first define each single statement in pseudo-code $\mathcal{R}$ (cf.\ Section~\ref{section:decomposing_complicated_task_into_rules}) as $p \rightarrow q$. 
Then, we check each statement sequentially:
for the current statement $p \rightarrow q$, the condition $p$ is checked if it is satisfied;
if $p$ is fulfilled, terminate the process and set the answer $ans = q$;
if not, proceed to the next statement;
continue this process until an answer is drawn (see the reasoning construction algorithm in Appendix~\ref{appendix-section:methodology}).
Next, the obtained $C_{\text{RSNG}}$ and $ans$ will be inserted into prompt. 
This process is repeated for all demonstrations to obtain the full set of reasoning steps. Notably, the pseudo-code can be extended for complex cases with multiple reasoning paths by adding an "ELSE" clause for alternative statements.

\begin{table*}[tbp!]
\centering
\footnotesize
\renewcommand{\arraystretch}{0.7}
\begin{tabular}{r|ll|ll|ll}
\toprule
                          & \multicolumn{2}{c}{\textsf{MIMIC}} & \multicolumn{2}{c}{\textsf{Synthea}} & \multicolumn{2}{c}{\textsf{CMS}}  \\
\multirow{-2}{*}{Methods} & Acc        & F1          & Acc         & F1                   & Acc       & F1       \\ \midrule
\SMAT{} \quad \quad       & 0.9865     & 0.2020      & 0.9902      & 0.3850     & 0.9852    & 0.5000         \\ \midrule
\texttt{GLM-4-Flash}      & 0.9512     & 0.0369      & 0.9895      & 0.1143     & 0.9922    & 0.1667      \\
\rowcolor{blue!10} 
\quad +\model{}           & 0.9816$_{+3.04\%}$      & \textbf{0.1061}$_{+6.92\%}$      & 0.9852$_{+0.43\%}$      & \textbf{0.1538}$_{+3.95\%}$     & 0.9836$_{-0.86\%}$    & \textbf{0.2500}$_{+8.33\%}$      \\ 
\midrule
\texttt{Llama-3-8B}       & 0.3272     & 0.0060      & 0.8870      & 0.0233     & 0.9926    & 0.0952      \\
\rowcolor{blue!10} 
\quad +\model{}           & 0.9774$_{+65.02\%}$     & \textbf{0.0765}$_{+7.05\%}$      & 0.9572$_{+7.02\%}$      & \textbf{0.0863}$_{+6.30\%}$     & 0.9481$_{-4.45\%}$    & \textbf{0.1192}$_{+2.40\%}$      \\ 
\midrule
\texttt{Mistral-7B}       & 0.9931     & 0.1538      & 0.9686      & 0.0792     & 0.9879    & 0.2051             \\
\rowcolor{blue!10} 
\quad +\model{}           & 0.9947$_{+0.16\%}$     & \textbf{0.3200}$_{+16.62\%}$      & 0.9895$_{+2.09\%}$      & \textbf{0.1143}$_{+3.51\%}$     & 0.9867$_{-0.12\%}$    & \textbf{0.2273}$_{+2.22\%}$       \\ 
\midrule
\texttt{GPT-3.5}          & 0.9767     & 0.0745      & 0.9865      & 0.0909     & 0.7656    & 0.0196           \\
\rowcolor{blue!10} 
\quad +\model{}           & 0.9944$_{+1.77\%}$     & \textbf{0.6207}$_{+54.62\%}$ & 0.9936$_{+0.71\%}$  & \textbf{0.4242}$_{+33.33\%}$    & 0.9867$_{+22.11\%}$      & \textbf{0.2917}$_{+27.21\%}$ \\ 
\midrule
\texttt{GPT-4o}           & 0.9964     & 0.3030 & 0.9919  & 0.0769 & 0.9922 & 0.1667  \\
\rowcolor{blue!10} 
\quad +\model{}           & 0.9959$_{-0.05\%}$     & \textbf{0.3500}$_{+4.70\%}$    & 0.9946$_{+0.27\%}$       & \textbf{0.4667}$_{+38.98\%}$  & 0.9887$_{+0.20\%}$    & \textbf{0.4082}$_{+24.15\%}$    \\ 
\bottomrule
\end{tabular}
\caption{\sld{The results for the schema matching task.}}
\label{table:main_aio}
\end{table*}

% \begin{figure}[htbp]
%     \centerline{\includegraphics[width=\columnwidth]{figures/SIE.pdf}}
%     \caption{Self-Indicator Extraction.}
%     \label{figure:self_indicator_extraction}
% \end{figure}

\paragraph{Self-Indicator Extraction}
Inspired by the performance gain observed when generating a high-quality summary at the beginning of reasoning in our experiments, we further add a Self-Indicator Extraction pre-task.
This plugin module generates a filtered text segment, called self-indicator ($K_{\text{SI}}$), describing key information from $\{r, r'\}$. 
Specifically, taking $\{r, r'\}$ and $K$ as input, an LLM is used to generate $K_{\text{SI}}$ (cf.\ Appendix~\ref{appendix-section:implementation}), which is then appended to the prompts created above.

\paragraph{Summarized Demonstrations}
After decomposing task instructions into pseudo-code, we leverage in-context learning to enable the LLM to learn reasoning behaviors beyond the concept of ``\texttt{match}''.
% Since the task instruction has been decomposed into pseudo-code in Section~\ref{section:decomposing_complicated_task_into_rules}, we leverage in-context learning to help LLM learn the reasoning behavior rather than modeling the concept of ``match''. 
% To this end, we again apply an LLM to summarize all the $\{d,d'\}$ in the demonstrations, for the sake of prompt length controlling. $\{d,d'\}$ in prompts will then be replaced by the summarized ones. 
To manage the prompt length, we use an LLM to summarize all demonstration pairs $\{d, d'\}$ (cf.\ Appendix~\ref{appendix-section:implementation}), %These summarization pairs replace the original $\{d, d′\}$ in the prompts, ensuring efficiency without losing essential information.
% Instead of using the original demonstrations $\{d, d'\}$, we 
and use \emph{summarized} pairs to improve efficiency without losing essential information.

\subsection{Inconsistency-tolerant Generation Ensemble (IntGE)}

With multiple knowledge sets available, a straightforward utilization is to combine all retrieved knowledge within one prompt, but this practice leads to information within a prompt flooded thus bringing unexpected outputs~\citep{parciak_schema_2024}. Therefore, we instead propose IntGE to integrate diverse information sources and improve output stability in a \emph{fine-tuning-free} manner.
We define the available knowledge sets from unique sources as $\mathcal{K}=[K_1, K_2, \ldots, K_n]$. 
For a given pair $\{r, r'\}$, we retrieve information from each source in $\mathcal{K}$, resulting in $n$ knowledge lists.
Each list is used to construct a prompt following the method in Section~\ref{section:prompt_construction}, creating $n$ prompts. 
We ensemble the $n$ binary classification results using majority voting to determine the prediction for $\{r, r'\}$.
% These prompts are sent to LLM to generate $N$ outputs. For $N$ binary classification results from outputs, we ensemble them using majority voting and take the voting result as the prediction of $\{r, r'\}$.

As noted in previous studies~\citep{cuconasu2024power, zhang_knowledge_2024}, LLMs tend to prefer familiar input from their pre-training phase, and distracting context can lead to poor outputs. IntGE aims to prevent interference between different knowledge sources by separating them into distinct prompts. Inspired by~\citet{wang_self-consistency_2023, shi2024replug}, we adopt a straightforward voting strategy. This approach leverages multiple knowledge sources while enhancing the model's robustness against poorly formatted outputs by keeping prompt lengths manageable.
% the motivation of IntGE is to avoid knowledge from various sources overshadowing each other by splitting them into different prompts, and driven by the observation that longer prompts lead to less decisive LLM outputs also the inspiration from~\citep{wang_self-consistency_2023,shi2024replug}, a simple enough voting strategy is adopted to utilize multiple knowledge sources while enhancing \model{}'s robustness towards badly-formatted outputs by limiting the length of a single prompt.

\section{Experiments}

\subsection{Experimental Setup}

\paragraph{Dataset and Metrics}
\sld{Following \citet{narayan2022can}, we choose the challenging \textsf{Synthea}, \textsf{MIMIC}, and \textsf{CMS} from the OMAP benchmark \citep{bellatreche_smat_2021} for SM. 
For EM, we adapt the MedMentions benchmark~\citep{mohan_medmentions_2019} to construct a new biomedical dataset \textsf{M}ed\textsf{M}entions \textsf{M}atching (\textsf{MMM}), with details in Appendix~\ref{appendix-section:implementation}. 
To assess \model{}'s generalization beyond the biomedical domain, we include widely used out-of-domain EM datasets: Amazon-Google (\textsf{AG}), Walmart-Amazon (\textsf{WA}), DBLP-Scholar (\textsf{DS}) from DeepM~\citep{mudgal_deep_2018}.}

% Based on the datasets above, we give the statistics as shown in Table 2, whereby it is clear that each dataset suffers from sample imbalance. Here, we further extend this issue by highlighting the main challenge of the matching task as the conflict between the practical need for higher recall and the sample-imbalanced real-world scenario.

% \paragraph{Metrics}
\added{Following \citet{bellatreche_smat_2021,narayan2022can}, we report F1-score across all pairs for its suitability in the \emph{inherently imbalanced} data matching tasks and include accuracy to show precision and recall trends after applying \model{}. Detailed precision and recall results are in Appendix~\ref{appendix-section:performance_analysis}.}
% across all pairs due to matching tasks' imbalanced nature.
% We also include accuracy alongside F1 to outline the trend of precision and recall after applying \model{} in this imbalanced scenario. 
% Detailed precision and recall results are provided in Appendix~\ref{appendix-section:performance_analysis}.

\paragraph{Settings}
We mainly follow Section~\ref{section:matching_framework} to conduct experiments.
However, \added{for \textsf{AG}, \textsf{WA} and \textsf{DS} in EM, those derived from e-commerce and publication-profile data lacking readily available knowledge bases, we test \model{} \emph{without} knowledge construction \& retrieval}. 
Unless otherwise noted, LLMs utilized in \model{} are \texttt{GPT-3.5-turbo-1106}. 
Detailed settings and all prompts, including pseudo-code, are in Appendix~\ref{appendix-section:methodology} and~\ref{appendix-section:prompts}. 
\sld{Given \model{}'s fine-tuning-free nature, we mainly compare it with (1) five \textbf{\emph{few-shot LLM baselines}} including \emph{API-styled} \texttt{GPT-3.5-turbo-1106}, \texttt{GPT-4o}, and \texttt{GLM-4-Flash} and \emph{locally-deployed} \texttt{Llama-3-8B-Instruct} and \texttt{Mistral-7B-Instruct}; and (2) \textbf{\emph{non-LLM competitors}}, namely \SMAT{} \citep{bellatreche_smat_2021} achieving the best SM performance on OMAP~(by classifying schema pairs just like \model{}) and \textsc{sudowoodo}~\citep{wang_sudowoodo_2023}, a self-supervised SoTA methods for EM tasks.}
% but also include results from . Specifically, we use  as LLM baselines.
% %
% \sld{We also include \SMAT{} \citep{bellatreche_smat_2021}, the best non-LLM method on OMAP, as a non-LLM baseline. 
% Since \SMAT{} is specifically designed for SM, we also include the self-supervised method \textsc{sudowoodo}~\citep{wang_sudowoodo_2023} for EM}.

\begin{table*}[tbp!]
\centering
\footnotesize
\renewcommand{\arraystretch}{0.55}
\begin{tabular}{r|cc|cc|cc|cc}
\toprule
\rowcolor{newtab!10}
                          & \multicolumn{2}{c}{\textsf{MMM}} & \multicolumn{2}{c}{\textsf{AG}} & \multicolumn{2}{c}{\textsf{WA}} & \multicolumn{2}{c}{\textsf{DS}} \\
\rowcolor{newtab!10}
\multirow{-2}{*}{Methods}       & Acc       & F1       & Acc        & F1                  & Acc         & F1                   & Acc       & F1                 \\ \midrule
\rowcolor{newtab!10}
\textsc{sudowoodo}              & -         & -       & 0.9492    & 0.5045   & 0.9532              & 0.5236              & 0.9089                   & 0.7706 \\
\midrule
\rowcolor{newtab!10}
\texttt{GLM-4-Flash}            & 0.9663    & 0.6841  & 0.6140             & 0.3410              & 0.4256              & 0.2332              & 0.9195              & 0.8009 \\
\rowcolor{newtabemph!20} 
\quad +\model{}                & 0.9690      & \textbf{0.7334}    & 0.8203              & \textbf{0.5164}     & 0.8404              & \textbf{0.5038}     & 0.9646              & \textbf{0.9035} \\
\rowcolor{newtab!10}
$\Delta$                       & $+0.27\%$   & $+4.92\%$    & $+20.63\%$          & $+17.54\%$          & $+41.48\%$          & $+27.06\%$          & $+4.51\%$           & $+10.26\%$ \\
\midrule
\rowcolor{newtab!10}
\texttt{Llama-3-8B}            & 0.9574   & 0.6738    & 0.7553             & 0.2761              & 0.2660              & 0.1861              & 0.8377              & 0.6799 \\
\rowcolor{newtabemph!20} 
\quad +\model{}                & 0.9730   & \textbf{0.7783}    & 0.6594              & \textbf{0.3561}     & 0.8023              & \textbf{0.4692}     & 0.9093              & \textbf{0.7637} \\
\rowcolor{newtab!10}
$\Delta$                       & $+1.56\%$   & $+10.45\%$    & $-9.59\%$           & $+8.00\%$           & $+53.63\%$          & $+28.31\%$          & $+7.16\%$           & $+8.38\%$ \\
\midrule
\rowcolor{newtab!10}
\texttt{Mistral-7B}            & 0.9671    & 0.7046    & 0.7479             & 0.4289              & 0.4300              & 0.2275              & 0.8555              & 0.6318 \\
\rowcolor{newtabemph!20} 
\quad +\model{}                & 0.9689    & \textbf{0.7206}   & 0.7771              & \textbf{0.4511}     & 0.8824              & \textbf{0.5594}     & 0.8948              & \textbf{0.6595}    \\
\rowcolor{newtab!10}
$\Delta$                       & $+0.18\%$   & $+1.60\%$    & $+2.92\%$           & $+2.22\%$           & $+45.24\%$          & $+33.19\%$          & $+3.93\%$           & $+2.77\%$ \\
\midrule
\rowcolor{newtab!10}
\texttt{GPT-3.5}               & 0.9297    & 0.6242    & 0.8731             & 0.2988              & 0.8380              & 0.4071              & 0.8915              & 0.6245 \\
\rowcolor{newtabemph!20} 
\quad +\model{}                & 0.9727        & \textbf{0.7758}     & 0.8809              & \textbf{0.5991}     & 0.9517              & \textbf{0.7288}     & 0.9300              & \textbf{0.8260}$_{\%}$  \\
\rowcolor{newtab!10}
$\Delta$                       & $+4.30\%$   & $+15.16\%$     & $+0.78\%$           & $+30.03\%$          & $+11.37\%$          & $+32.17\%$          & $+3.85\%$           & $+20.15\%$ \\
\midrule
\rowcolor{newtab!10}
\texttt{GPT-4o}                & 0.9658 & 0.6738     & 0.9097             & 0.6810              & 0.9688              & 0.8469              & 0.9613              & 0.8920 \\
\rowcolor{newtabemph!20} 
\quad +\model{}                & 0.9661   & \textbf{0.6744}     & 0.9128              & \textbf{0.6875}     & 0.9746              & \textbf{0.8750}     & 0.9653              & \textbf{0.9035} \\
\rowcolor{newtab!10}
$\Delta$                       & $+0.03\%$   & $+0.06\%$     & $+0.31\%$           & $+0.65\%$           & $+0.58\%$           & $+2.81\%$           & $+0.40\%$           & $+1.15\%$ \\
\bottomrule
\end{tabular}
\caption{\textcolor{newtabcap}{The results for the entity matching task.}}
\label{table:main_aio_em}
\end{table*}
 
\subsection{Main Results for SM and EM}

\paragraph{\emph{\model{} achieves superior performance in SM}}
As shown in Table~\ref{table:main_aio}, generally \model{} is significantly better than all LLM baselines and \SMAT{} for SM. 
\added{Though this superiority is less pronounced on \textsf{CMS}, a notable performance improvement over LLM baselines is still evident}. 
Specifically, on \textsf{Synthea}, \texttt{GPT-3.5}-version \model{} outperforms the previous \SMAT{} by 3.92\% in F1-score, and on \textsf{MIMIC}, the gain increases to 41.87\%. In addition, we observe that the F1-score of \texttt{GPT-3.5}-version \model{} outperforms all baselines by an average of 33.73\% and 50.59\% on \textsf{Synthea} and \textsf{MIMIC}, respectively. On \textsf{MMM}, the F1-score of \model{} surpasses all LLM baselines by an average of 10.62\%, \added{and on \textsf{CMS} this gain reaches 12.86\%}.

\paragraph{\emph{\model{} ensures positive sample recall in SM}}
\sld{Although we see a slight drop in accuracy for \texttt{GPT-4o}-version \model{}, this 0.05\% drop is not caused by a failure in recall, as detailed precision and recall in Appendix~\ref{appendix-section:performance_analysis} show.
Instead, it results from an increase in both true-positive and false-positive recalls, specifically, the drop is due to the correct recalling of 2 extra positive samples and incorrect recalling of 5 negative samples, which looks reasonable for the 2 more true-positive raise the recall from 38.5\% to 53.8\% while keeping the precision around 25.0\%, so we still consider this a strong result for an inherently imbalanced matching task.}

\paragraph{\emph{\model{} generalizes well on out-of-domain EM}}
\added{As shown in Table~\ref{table:main_aio_em}, \model{} consistently enhances the performance of all LLMs for EM, aligning with the SM results in Table~\ref{table:main_aio}.
Even without knowledge retrieval, \model{} remains competitive with the semi-supervised method \textsc{sudowoodo}, highlighting its strong generalizabilities across commonly used EM datasets.}

\paragraph{\emph{\model{} generalizes across backbones for both SM and EM}}
We observe that after implementing \model{}, \sld{the F1-score of each backbone is improved by an average of 17.98\%, 17.21\%, 8.91\%, 6.44\%, 11.69\%, 24.71\%, 8.54\% on \textsf{MIMIC}, \textsf{Synthea}, \textsf{CMS}, \textsf{MMM}, \textsf{AG}, \textsf{WA}, and \textsf{DS}, respectively}. 
These results demonstrate that our proposed \model{} can generalize across different backbones and consistently improve the performance of LLMs on both SM and EM tasks. 
Notably, on \textsf{MIMIC}, the strongest version of \model{} achieves an F1-score of 0.6207, clearly outperforming a fine-tuned SoTA \Jellyfish{}~\citep{zhang_jellyfish_2024} for SM, which achieves 0.4314. This result indicates the great potential of fine-tuning-free LLMs in data matching tasks on domain-specific data.

\subsection{Component Analysis}

\paragraph{\emph{Task-specific pseudo-code disambiguates task instructions}}
The pseudo-code is designed to prevent LLM confusion about task objectives by providing explicit conditional statements.
To explore its effectiveness, we examine \model{} on \textsf{Synthea} without our proposal: all pseudo-code and corresponding reasoning steps are removed from the demonstrations, and the results are compared with the complete \model{}. 
We classify false-positive errors into three types: \textit{Over-matching} (OM), as described in Figure~\ref{figure:matching_errors}; \textit{Position Mismatching} (PM), where parts of $r$ and $r'$ (e.g., a schema's table and another's column) are incorrectly matched; and \textit{Incorrect Reasoning} (IR), covering errors not fitting the first two categories. Table~\ref{table:analysis_pseudo_code} shows that the pseudo-code reduces \emph{all} three error types. %more specifically, it reduces the number of over-matched cases by 47, incorrect-reasoned cases by 3 and position-mismatched cases by 10. 

\begin{table}[htbp!]
\centering
\footnotesize
\renewcommand{\arraystretch}{0.7}
\begin{tabular}{cccc}
\toprule
\multirow{2}{*}{Settings} & \multicolumn{3}{c}{\#~(Case)} \\
  & IR & OM & PM \\
\midrule
w/o pseudo-code   & 8 & 60 & 11 \\
w. pseudo-code& 5 & 13 & 1  \\
\midrule
$\Delta$  &-37.5\% &-78.3\% &-90.9\%   \\
\bottomrule
\end{tabular}
\caption{Number of erroneous cases with and without the proposed pseudo-code on \textsf{Synthea}.}
\label{table:analysis_pseudo_code}
\end{table}

\begin{figure}[htbp]
    \centerline{\includegraphics[width=\columnwidth]{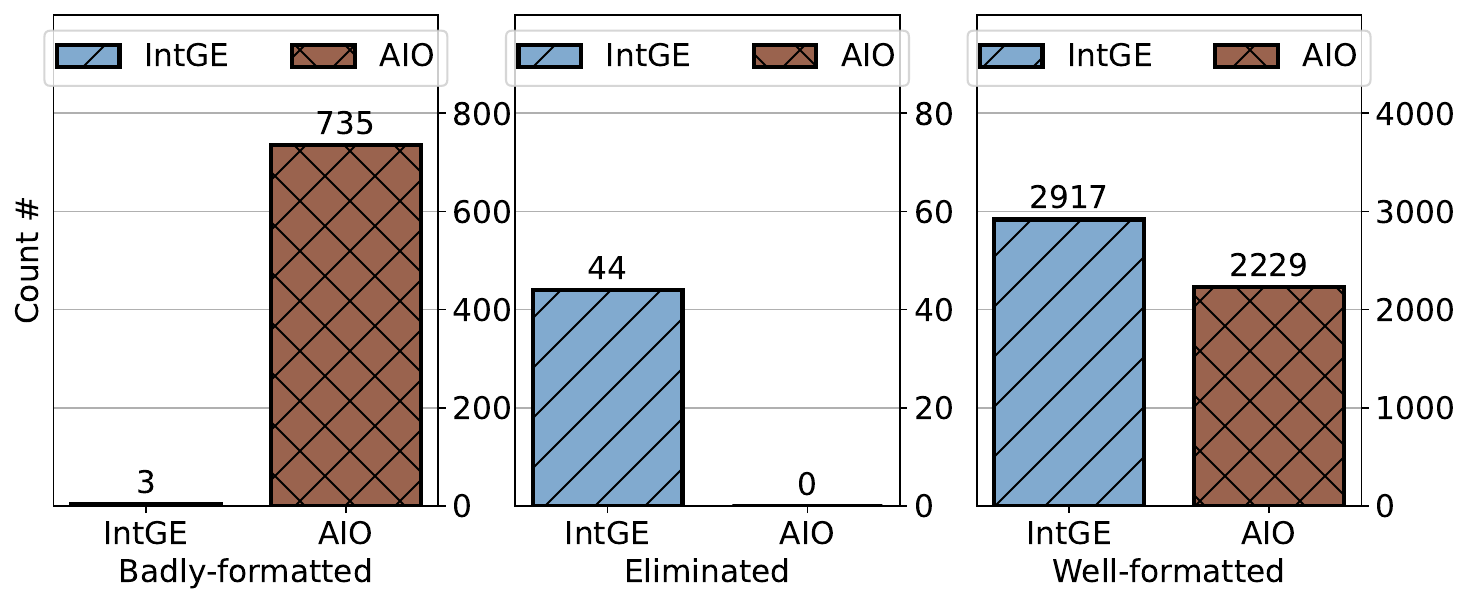}}
    \caption{Ablations of IntGE on \textsf{Synthea}.}
    \label{figure:analysis_intge}
\end{figure}

\paragraph{\emph{IntGE enhances robustness against badly-formatted outputs}}
To assess the impact of IntGE, we test a \emph{knowledge-all-in-one} (AIO) version of \model{}, which removes IntGE and combines knowledge from all sources into a single prompt.
%
% The results are categorized into three types as shown in Figure~\ref{figure:analysis_intge}. \emph{Badly-formatted} means the output format of the \model{} does not follow the format given in the demonstrations, and \emph{Well-formatted} means the opposite. \emph{Eliminated} means that the badly-formatted output appeared in IntGE's votes but was eventually eliminated by IntGE's voting strategy, thus the final output remains Well-formatted. 
As shown in Figure~\ref{figure:analysis_intge}, results are categorized into three types: \emph{Badly-formatted}, where the output does not follow the demonstrated format; \emph{Well-formatted}, where it does; and \emph{Eliminated}, where IntGE's voting mechanism removes poorly formatted outputs, ensuring the final result is well-formatted.
The results clearly indicate that the AIO setting performed poorly. In contrast,
% According to Figure~\ref{figure:analysis_intge} we observe that the outputs completely go wild under the knowledge-all-in-one setting, while 
IntGE effectively reduces the number of badly-formatted output via its multi-prompt voting mechanism. The results imply a paradigm for retrieving knowledge from different sources for inference-only LLMs, i.e., splitting retrieved knowledge into separate prompts based on the source to improve output quality while managing prompt length.

% \begin{figure}[ht]
% \centering
% \begin{subfigure}[b]{0.5\columnwidth}
%  \centering
%  \includegraphics[width=\columnwidth]{figures/analysis_illegal_IntGE_wocb.pdf}
%  \caption{IntGE}
%  \label{subfigure:analysis_intge_intge}
% \end{subfigure}%
% \hfill
% \begin{subfigure}[b]{0.5\columnwidth}
%  \centering
%  \includegraphics[width=\columnwidth]{figures/analysis_illegal_AIO_wocb.pdf}
%  \caption{All in One}
%  \label{subfigure:analysis_intge_aio}
% \end{subfigure}
% \caption{Visualization of the results on \textsf{Synthea} with and without IntGE: (a) is the vanilla \model{} with IntGE, (b) is the knowledge-all-in-one version of \model{} without IntGE. The deepest color indicates that \model{} outputs a badly-formatted response on the example.}
% \label{figure:analysis_intge}
% \end{figure}

\added{In addition to the above analysis, detailed ablation results in Appendix~\ref{appendix-section:ablation} and~\ref{appendix-section:cost_analysis} further confirm the necessity and effectiveness of each component of \model{}.}

% Based on this observation, grouping retrieved knowledge into different prompts by their sources to limit the length of the prompt is a feasible enough way to ensure decisiveness while using inference-only LLMs.

\section{Conclusion}

We presented \model{}, a fine-tuning-free framework for solving data matching tasks under a knowledge-enhanced setting, with a pseudo-code based task instruction strategy for addressing the problem of confusion, and mechanisms DaK, EaK, and IntGE for building and utilizing knowledge from various sources in scarcity of unstructured domain knowledge. \sld{Our evaluations show that \model{} enhances five LLM backbones and outperforms the current non-LLM solutions in most cases in a completely fine-tuning-free paradigm, and our pseudo-code effectively mitigated LLMs' confusion towards the matching tasks.} %We also showed that limiting the length of a single prompt led to mitigate badly-formatted outputs when tackling knowledge-enhanced matching tasks.
Our future work involves extending \model{} to make it suitable for both metadata-based and instance-based scenarios, and further improving the efficiency of the matching workflow.

\clearpage
\section{Limitations}

One limitation of our approach is the reliance on human effort to select appropriate domain knowledge bases (KBs) as sources of information, which may pose challenges for practical implementation.
Also, this study primarily focuses on matching tasks framed as predicting enumerated data pairs rather than directly identifying potential mappings within source databases. While this approach simplifies the problem formulation, it may lead to increased computational demands and thus have an environmental impact. 
\added{Although cost is a concern, our cose analysis in Appendix~\ref{appendix-section:cost_analysis} shows the cost-in-total remains acceptable. Nontheless, we recognize the unexplored} opportunity for efficiency optimization in future work.

% Bibliography entries for the entire Anthology, followed by custom entries
%\bibliography{anthology,custom}
% Custom bibliography entries only
\bibliography{custom}

\clearpage
% \onecolumn
\appendix

\section{Methodology}
\label{appendix-section:methodology}

\added{Appendix~\ref{appendix:example} provides examples illustrating the key concepts from the main text used to construct the final prompt.
Appendix~\ref{appendix:entity_matching} discusses the adaptations required for applying \model{} to entity matching tasks.
Appendix~\ref{appendix:algorithms} presents algorithms formalized based on the main procedure described in Section~\ref{section:matching_framework}.}

\added{\subsection{Examples}\label{appendix:example}}

\added{The following examples correspond to the \model{} techniques described in Section~\ref{section:matching_framework} and illustrated in Figure~\ref{figure:main}.}

\begin{figure}[!htbp]
\centerline{\includegraphics[width=\columnwidth]{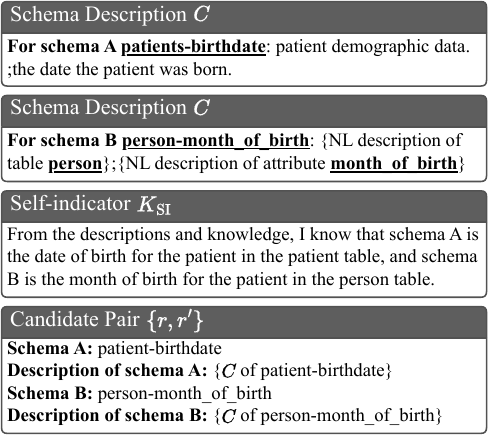}}
\caption{Examples of schema description $C$, self-indicator $K_\text{SI}$, and candidate pair $\{r,r'\}$ based on the schema \emph{patients-birthdate} and \emph{person-month\textunderscore of\textunderscore birth}.}
\label{figure:examples}
\end{figure}

\begin{figure}[!htbp]
\centerline{\includegraphics[width=\columnwidth]{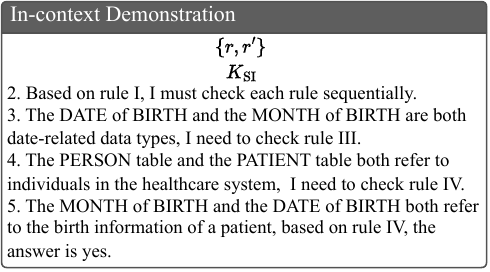}}
\caption{An example of the in-context learning demonstration, i.e., an example of the [\emph{Examples}] in Table~\ref{appendix-table:kcmf_sm}, directly derived from the components from Figure~\ref{figure:examples}.}
\label{figure:example_icl}
\end{figure}

\subsection{Discussions on Entity Matching (EM)}
\label{appendix:entity_matching}

Since entity matching operates at the level of records, the task of entity matching can be defined by slightly modifying the definition of $r$ described in Section~\ref{section:problem_statement} for schema matching: 
\begin{equation}			
\label{equation:entity_mappings}
r = \{N, {attr}_1, {attr}_2, \ldots, {attr}_m\},
\end{equation}
where $N$ represents the name of the entity $r$, but the schema description $C$ is replaced with a sequence of attributes $\{ {attr}_1, {attr}_2, \ldots, {attr}_m \}$, which corresponds to all the other column values of the entity $r$.

\added{For \model{}, the only difference between solving schema matching (SM) and entity matching (EM) tasks is the number of attributes requiring serialization.
In SM, \model{} only needs to serialize schema name and description, whereas in EM, \model{} needs to serialize the entity name and potentially a variable number of attributes. 
In addition, since SM operates at the schema level while EM at the record level, EM datasets are often much larger in scale in practice.
As a result, introducing \emph{blocking} as a preprocessing phase, as commonly performed in conventional EM methods~\cite{ebraheem_deeper_2018,zhang2020autoblock}, could achieve higher efficiency in handling those larger EM datasets.}

% consideration of privacy can be no longer fulfilled by using metadata instead because the record is indispensable.

\subsection{Algorithms}
\label{appendix:algorithms}

\added{Algorithm~\ref{algorithm:cross_example_information_discovery_algorithm} formalizes from the metadata discovering procedure introduced in Section~\ref{section:domain_knowledge_set}.
As discussed in Section~\ref{section:domain_knowledge_set}, the functions $\operatorname{ObjectDiscovery}()$ and $\operatorname{FindMetadata}()$ can be implemented using readily available Named Entity Recognition (NER) models or LLMs.}

\begin{algorithm}[!htbp]
\caption{Metadata Discovery}
\label{algorithm:cross_example_information_discovery_algorithm}
\footnotesize
\begin{flushleft}
\textbf{Input}: A matching task dataset $D$ \\
\textbf{Output}: A metadata list $\mathcal{C}_{\text{DaK}}$ and an object list $\mathcal{C}_{\text{Obj}}$
\end{flushleft}
\begin{algorithmic}[1] %[1] enables line numbers
\STATE $\mathcal{C}_{\text{DaK}} \gets \emptyset$, $\mathcal{C}_{\text{Obj}} \gets \emptyset$
\FOR{$\{r_i, r_i'\} \in D$}
\STATE $O \gets \operatorname{ObjectDiscovery}(r_i, r'_i)$ 
\FOR{{$\{r_j, r_j'\} \in D, j \neq i$}}
\STATE $K_{\text{DaK}} \gets \operatorname{FindMetadata}(O, r_j, r'_j)$
\IF{$K_{\text{DaK}} \neq \text{NULL}$}
\STATE $\mathcal{C}_{\text{DaK}}.insert(K_{\text{DaK}})$
\STATE $\mathcal{C}_{\text{Obj}}.insert(O)$
\ENDIF
\ENDFOR
\ENDFOR
\STATE \textbf{return} $\mathcal{C_{\text{DaK}}}$, $\mathcal{C_{\text{Obj}}}$
\end{algorithmic}
\end{algorithm}

\added{Algorithm~\ref{algorithm:rule_based_reasoning_prompting} formalizes the scan-then-check procedure over statements while constructing prompts, which has been discussed in Section~\ref{section:prompt_construction}.}

\begin{algorithm}[!htbp]
\caption{Constructing Reasoning Steps}
\label{algorithm:rule_based_reasoning_prompting}
\footnotesize
\begin{flushleft}
\textbf{Input}: Current demonstration $\{d, d'\}$ and the pseudo-code $\mathcal{R}$ \\
\textbf{Output}: A reasoning steps $C_{\text{RSNG}}$ and a corresponding final answer $ans$ 
\end{flushleft}
\begin{algorithmic}[1] %[1] enables line numbers
\STATE $C_{\text{RSNG}} \gets \emptyset$
\FOR{$R \in \mathcal{R'}$}
\STATE $C_{\text{RSNG}}.append(\operatorname{genReasoningStep}(R, {d, d'}))$
\IF{$\operatorname{isConditionMet}(R, {d, d'})$ or $\operatorname{isLastStatement}(R, \mathcal{R})$}
\STATE $ans \gets \operatorname{getQ}(R)$
\STATE \textbf{return} $C_{\text{RSNG}}$, $ans$
\ENDIF
\ENDFOR
\end{algorithmic}
\end{algorithm}

\added{Here, in \Cref{algorithm:rule_based_reasoning_prompting}, $\operatorname{getQ}()$ takes a statement $R$ as input and returns its conclusion~(the $q$ from $p\rightarrow q$).}

\section{Implementation}
\label{appendix-section:implementation}

\added{In this section, we present detailed information on implementing \model{}, specifically, 
B1. guidelines for designing the pseudo-code for a specific task~(i.e., SM or EM), 
B2. the knowledge sources and the models used for knowledge retrieval and construction, 
B3. the models and parameters used for pre-tasks~(e.g., self-indicator extraction) during prompt construction, 
B4. the models and parameters used in inference, 
B5. the modifications made to the MedMentions dataset to construct \textsf{MMM}, and 
B6. the accessibility datasets and code used in the experiments.}

\paragraph{B1. Pseudo-code Design}
\label{appendix:implementation_pseudo_code_design}
The pseudo-code used in our experiments for SM and EM is shown in Table~\ref{appendix-table:pseudo_code_sm} and Table~\ref{appendix-table:pseudo_code_em}, respectively.
It was directly authored based on the objectives of the matching tasks.

As emphasized in the main text, the pseudo-code statements are defined as ``IF-THEN-ELSE''. 
To better utilize the LLM's natural language comprehension capability, this structure is adapted to ``IF, THEN, otherwise'' in the actual design of the pseudo-code.
To ensure that the given pseudo-code is checked in the correct order, a statement is added at the beginning of the designed pseudo-code to explicitly specify that the subsequent statements should be checked in order. This precautionary step can be omitted when using more advanced LLMs (e.g., \texttt{GPT-4o}).

Additionally, to cover all the task motivations of the matching task, we include overlapped conditions in the pseudo-code statement design. 
For the last statement of all the pseudo-codes, we set a general condition to guarantee comprehensive coverage of all cases.

\paragraph{B2. Knowledge Retrieval \& Construction}
\label{appendix:implementation_knowledge_retrieval_construction}
The retrieval pipeline in this paper is based on keywords extracted from $\{r,r'\}$, then the quality-managed keywords are used to retrieve and construct the knowledge sets following the techniques demonstrated in Section~\ref{section:matching_framework}. Specifically, we adopt \texttt{GPT-3.5} to extract domain-specific, difficult-to-understand keywords from $\{r,r'\}$ (see Table~\ref{appendix-table:keywords_extraction} for details of the prompt). After that, we use \texttt{GPT-3.5} to filter the extracted keywords (see Table~\ref{appendix-table:keywords_quality_sm} and Table~\ref{appendix-table:keywords_quality_em} for details of the prompt), and we empirically design a blacklist for further rejecting low-quality keywords. 

To construct the EaK knowledge set, we select \textsc{Snomed-CT} as the knowledge base (KB) and use its API\footnote{\url{https://browser.ihtsdotools.org/snowstorm/snomed-ct}} based on the above keywords to search for the associated entities in the KB, specifically here we keep the top-$1$ search results, and for the searched entities, we query their children entities again using the API and randomly sample up to 3 children entities. The parent and children are then serialized into EaK knowledge in the form of ``One of {parent} is {children}’’.

For out-of-domain knowledge sources, we based on the English \textsc{Wikidata} API\footnote{\url{https://www.wikidata.org/w/api.php}} to search for entity codes using extracted keywords. We use the searched top-$1$ entity code to construct SPARQL statements for knowledge query from \textsc{Wikidata}\footnote{\url{https://query.wikidata.org/}} while using this entity code to query the page extraction from English \textsc{Wikipedia}\footnote{\url{https://en.wikipedia.org/w/api.php}}. Then, for the retrieved results, we use \texttt{GPT-3.5} to summarize the retrieved extractions to limit their length to no more than 1000 words under a zero-shot setting (see Table~\ref{appendix-table:retrieval_summarization}). 

\paragraph{B3. Prompt Construction}
\label{appendix:implementation_prompt_construction}
\texttt{GPT-3.5} is used to extract the self-indicator of schema pairs (or entity pairs in \textsf{MMM}).
The used prompts can be found in \ref{appendix-table:si_extraction}. To implement Demonstration-summarization-only, we also use \texttt{GPT-3.5} to compress the input $\{r,r'\}$. Specifically, to control the length of the prompt, we split $\{r,r'\}$ into two queries (see Table~\ref{appendix-table:dos} for the detailed prompt). 

\paragraph{B4. Inference Settings}
\label{appendix:implementation_inference_settings}
For the \textsf{MIMIC}, \textsf{Synthea} and \textsf{CMS} datasets, we take \textsc{Wikidata} with DaK, \textsc{Wikipedia} with EaK, and EaK as the knowledge sources to evaluate \model{} under the 4-shot setting. 
For \textsf{MMM}, we take \textsc{Wikidata}, \textsc{Wikipedia}, and EaK as the knowledge sources to evaluate the framework under the 2-shot setting. 
We empirically set temperature to 0 and top\textunderscore$p$ to 0.1 for stable outputs if these two parameters are available. Also, we directly took poorly formatted outputs as negative predictions in our experiments for metrics calculation and workflow automation. The pseudo-code we designed for these two tasks can be found in Appendix~\ref{appendix-section:prompts}. We derived the main results from the best result of running five times.

\begin{figure}[!htbp]
\centerline{\includegraphics[width=\columnwidth]{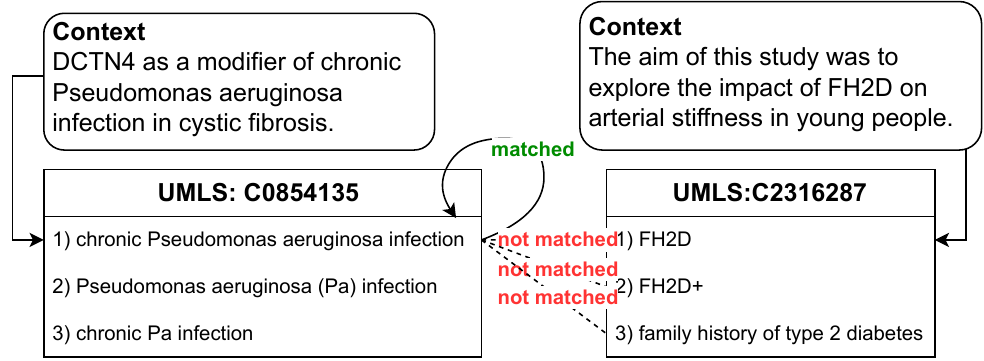}}
\caption{The process of constructing MedMentions Matching (\textsf{MMM}) dataset.}
\label{figure:MMM_construction}
\end{figure}

\begin{figure}[htb]
\centerline{\includegraphics[width=\columnwidth]{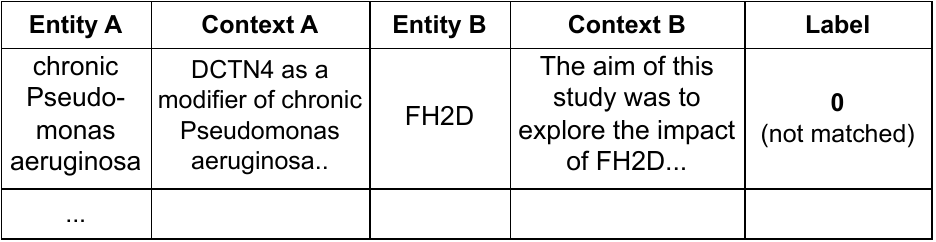}}
\caption{An example from \textsf{MMM}.}
\label{figure:MMM_example}
\end{figure}

\paragraph{B5. Construction of the \textsf{MMM} Dataset}
We present \textbf{M}ed\textbf{M}entions \textbf{M}atching, or \textsf{MMM}, a clinical entity matching dataset modified from MedMentions~\citep{mohan_medmentions_2019}. MedMentions is an entity linking dataset whose data are recognized entity mentions from PubMed documents labeled with Unified Medical Language System (UMLS) code. Inspired by the fact that entities referred to the same real-world concept are matched from the view of EM, we re-organized the MedMentions as per Figure~\ref{figure:MMM_construction}: For entities belonging to the same UMLS code, we keep the pairs with lowest similarity as positive. For those belonging to different, we sample the pairs with the highest similarity as negative. Then we use sentences where sampled entities are located as context. In total, \textsf{MMM} contains 7,359 positive and 100,000 negative examples and an example can be found in Figure~\ref{figure:MMM_example}.

\paragraph{B6. Dataset and Code}
All datasets can be accessed through original papers cited in this study, specifically for modifying \textsf{MMM}, the script is given in our supplementary codes. \added{Table~\ref{table:dataset_stat} presents the statistics of the test sets used in the main experiment, sampled as 10\% stratified subsets.}

\begin{table}[htbp!]
\centering
\footnotesize
\renewcommand{\arraystretch}{0.6}
\begin{tabular}{ccrrr}
\toprule
  Dataset & Task & \#~(Instance) & \#~(Positive) &  IR  \\
\midrule
\textsf{Synthea}         &   SM  &   2964    &   8   &   370.5   \\
\textsf{MIMIC}           &   SM  &   6408    &   13  &   492.9   \\

\textsf{CMS}             &   SM  &   2564    &   20   &   128.2   \\\midrule
\textsf{MMM}             &   EM  &   10736   &   736 &   14.6    \\
\textsf{AG}   &   EM  &   2293    &   234   &   9.8   \\
\textsf{WA}  &   EM  &   2049    &   193   &   10.6   \\
\textsf{DS}    &   EM  &   5842    &   1070   &   5.5   \\
\bottomrule
\end{tabular}
\caption{\sld{Statistics of evaluation datasets. IR = \#~(Instance) / \#~(Positive) is the Imbalance Ratio. The larger the IR, the harder the matching task.}} %the larger the dataset's imbalanced extent, and thus 
\label{table:dataset_stat}
\end{table}

\section{Performance Gain Analysis}
\label{appendix-section:performance_analysis}

\begin{table}[!htbp]
\centering
\footnotesize
\begin{adjustbox}{max width=\columnwidth}
\begin{tabular}{r|rrrrrr}
\toprule
\rowcolor{newtab!10}
                             & \multicolumn{2}{c}{\textsf{MIMIC}} & \multicolumn{2}{c}{\textsf{Synthea}} & \multicolumn{2}{c}{\textsf{CMS}} \\
\rowcolor{newtab!10}
\multirow{-2}{*}{Methods}    & P           & R           & P            & R            & P          & R          \\
\midrule
\rowcolor{newtab!10}
\SMAT{}                      & 11.5        & 84.6        & 24.4        & 90.9         & 33.9        & 95.0       \\
\midrule
\rowcolor{newtab!10}
\texttt{GLM-4}                          & 1.9         & 46.2        & 7.4         & 25.0          & 50.0        & 10.0      \\
\rowcolor{newtab!10}
+\model{}                    & 5.9         & 53.8        & 9.7         & 37.5          & 19.4        & 35.0      \\
\rowcolor{newtabemph!20} 
$\Delta$                     & $_{+4.0}$   & $_{+7.6}$   & $_{+2.3}$   & $_{+12.5}$    & $_{-30.6}$   & $_{+25.0}$ \\
\midrule
\rowcolor{newtab!10}
\texttt{Llama-3}                        & 0.3         & 100.0       & 1.2         & 50.0          & 100.0       & 5.0      \\
\rowcolor{newtab!10}
+\model{}                    & 4.2         & 46.2        & 4.6         & 75.0          & 6.9         & 45.0      \\
\rowcolor{newtabemph!20} 
$\Delta$                     & $_{+3.9}$   & $_{-53.8}$  & $_{+3.4}$   & $_{+25.0}$    & $_{-93.1}$  & $_{+40.0}$ \\
\midrule
\rowcolor{newtab!10}
\texttt{Mistral}                      & 10.3        & 30.8        & 4.3         & 50.0          & 21.1        & 20.0      \\ 
\rowcolor{newtab!10}
+\model{}                    & 21.6        & 61.5        & 9.1         & 37.5          & 20.8        & 25.0      \\
\rowcolor{newtabemph!20}
$\Delta$                     & $_{+11.3}$  & $_{+30.7}$  & $_{+4.8}$   & $_{-12.5}$   & $_{-0.3}$   & $_{+5.0}$ \\
\midrule
\rowcolor{newtab!10}
\texttt{GPT-3.5}                      & 4.1         & 46.2        & 5.6          & 25.0         & 1.0         & 30.0       \\ 
\rowcolor{newtab!10}
+\model{}                    & 56.3        & 69.2        & 28.0         & 87.5         & 25.0        & 35.0       \\
\rowcolor{newtabemph!20}
$\Delta$                    & $_{+52.2}$  & $_{+23.0}$  & $_{+22.4}$   &$_{+62.5}$    &$_{+24.0}$  &$_{+5.0}$  \\
\midrule
\rowcolor{newtab!10}
\texttt{GPT-4o}                       & 25.0        & 38.5        & 5.6          & 12.5         & 50.0       & 10.0       \\ 
\rowcolor{newtab!10}
+\model{}                    & 25.9        & 53.8        & 31.8         & 87.5         & 34.5       & 40.8       \\
\rowcolor{newtabemph!20}
$\Delta$                    & $_{+0.9}$   & $_{+15.3}$  & $_{+26.2}$   &$_{+75.0}$    &$_{-15.5}$  &$_{+29.2}$  \\
\bottomrule
\end{tabular}   
\end{adjustbox}
\caption{\textcolor{newtabcap}{Precision (P) and Recall (R) measures in SM. \model{} excels in imbalanced tasks like SM by effectively balancing precision and recall, leveraging pseudo-code to enhance recall without compromising precision.}}
\label{table:pr}
\end{table}

\added{To further explore how \model{} achieves a higher F1-score compared to both LLM baselines and the conventional method, we provide a detailed analysis of Precision (P) and Recall (R) on the three SM datasets with the highest imbalance ratios (see Table~\ref{table:dataset_stat}).
The results are provided in Table~\ref{table:pr}.
}

\added{In \emph{imbalanced} tasks like SM, often struggle to balance precision and recall, a challenge also evident in LLM baselines.
As shown in Table~\ref{table:pr}, different LLM baselines exhibit varying tendencies toward recall.
However, \model{} demonstrates performance improvements by addressing these imbalances: for baselines that prioritize precision, \model{} enhances recall by identifying more positive samples while effectively managing false positives, ensuring precision remains stable. This aligns with the practical preference for higher recall in matching tasks, reducing the need for additional human verification. 
For baselines that emphasize recall, \model{} improves precision by correctly rejecting negative samples, with less emphasis on further recall gains when unnecessary.
% despite which, we observe that the performance gain shown by \model{} is mainly achieved by maintaining the baseline's strength while trying to improve its weakness: for a baseline prefers precision, \model{} recalls more positive samples and successfully controls the growth of false-positive predictions so that the precision does not drop drastically, which exactly cater to the preference of matching tasks for recalls in practice, eliminating the need for additional human efforts to perform complex verification on recalled results; For a baseline prefers recall, \model{} correctly rejects negative samples, as its improvement in recalling becomes trivial in some cases.
}

\added{The performance gain of \model{} owes to its ability to enhance recall through pseudo-code, which operates independently of the model's reliance on the exposed data distribution.
This enables \model{} to accurately reject negative instances while improving the recall of positive samples.}

\section{Ablation Study}
\label{appendix-section:ablation}

\begin{figure}[htb]
\centerline{\includegraphics[width=\columnwidth]{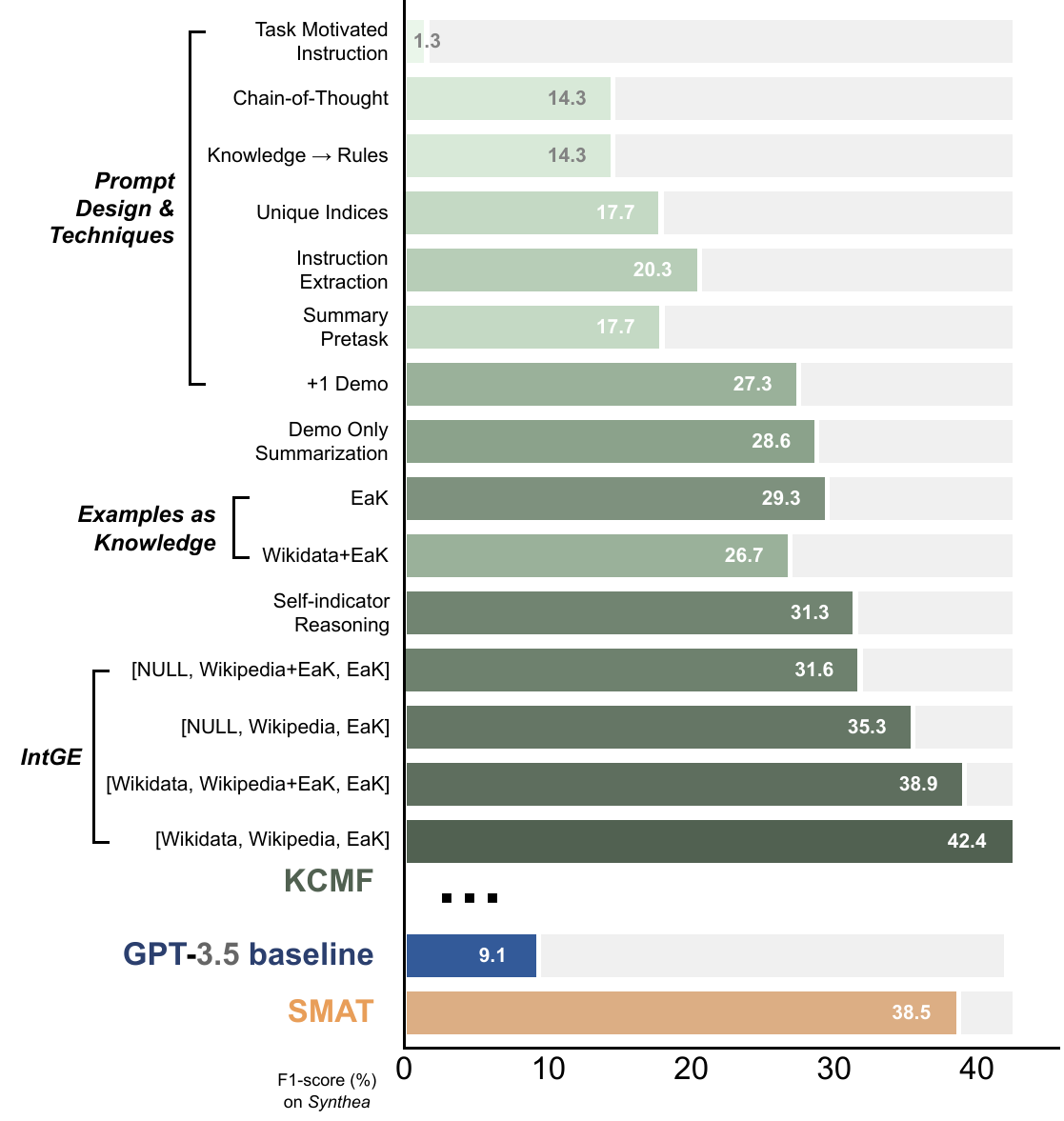}}
\caption{\model{} evolved from a plain zero-shot LLM query method. The horizontal axis depicts the F1-score for each version of \model{} on \textsf{Synthea}.}
\label{figure:ablation_results}
\end{figure}

\added{In this section, we incrementally stack techniques onto a bare LLM, demonstrating how techniques proposed in the main text transform a backbone model into a comprehensive matching framework.
The evolution process is visually summarized in Figure~\ref{figure:ablation_results} taking the SM task on \textsf{Synthea} as an example, which provides an overview of how each technique incrementally enhances the F1-score.
The process begins with a plain zero-shot LLM query (\texttt{GPT-3.5} baseline), which achieves an initial F1-score of 9.1\%. Through the addition of various techniques, the final framework, \model{}, attains an F1-score of 42.4\%, significantly outperforming both the \texttt{GPT-3.5} baseline (9.1\%) and the \SMAT{} method (38.5\%). The following subsections detail the techniques that contribute to this substantial performance improvement.
}

\subsection{Prompt Design and Construction}

We conducted incremental experiments using \textsc{Wikidata}+DaK with a 3-shot setting on \textsf{Synthea} as the baseline. 
The results, summarized in Table~\ref{table:ablation_tricks}, illustrate the impact of each optimization.

\begin{table}[!htbp]
\centering
\footnotesize
\begin{tabular}{r|rrr}
\toprule
Increment	 	&	Shot	&	Accuracy	&	F1-score \\
\midrule
TO Inst. 	& 3 & 0.5735 & 0.0125 \\ 
CoT 	& 3 & 0.9798 & 0.1429 \\ 
Know2Rule   	& 3 & 0.9798 & 0.1429 \\
U-indices 	& 3 & 0.9811 & 0.1765 \\
Inst. Extraction	& 3 & \underline{0.9841} & \underline{0.2034} \\
Summary Pretask 	& 3 & 0.9781 & 0.1772 \\
+1 demo	 	& 4 & \textbf{0.9892} & \textbf{0.2727} \\
\bottomrule
\end{tabular}
\caption{Ablation studies for prompt design and construction.}
\label{table:ablation_tricks}
\end{table}

\begin{itemize}[leftmargin=*]
\item \textbf{TO Inst.} Task-Oriented Instruction: We revised the instruction from \citet{narayan2022can} to ``Can records in schema B be transformed and stored into schema A?'' to better align with the schema matching task's objectives.

\item \textbf{Know2Rule} (Knowledge to Rule): The ``knowledge for the task'' prompt element was renamed to ``rules for the task'' for improved clarity and alignment with task requirements.

\item \textbf{U-indices} (Unique Indices): We applied unique indices for the different elements of the prompt that need to be numbered. We used Roman numerals, lowercase letter numbers, and numeric numbers separately for the row number of pseudo-code, the sequence number of retrieved knowledge, and sequence number of reasoning steps.

\item \textbf{Inst. Extraction} (Instruction Extraction): We extracted the instructions and rules for each demonstration and placed them at the beginning of the prompt.

\item \textbf{Summary Pretask} and \textbf{+1 Demo}: The Summary Pretask was designed to shorten prompt length, and +1 Demo added an extra demonstration. 
\end{itemize}
All the optimizations we added are motivated by achieving higher precision while ensuring promising recalls. 

As shown in Table \ref{table:ablation_tricks}, CoT, U-indices and Inst. Extraction brought 13.04\%, 3.36\%, 2.69\% improvement in F1-score on \textsf{Synthea}. Although Summary Pretask caused a temporary drop in performance, the combination of Summary Pretask and +1 Demo still shows a strong result; the F1-score is improved by 6.93\% compared to the Inst. Extraction step. 
We examine the results of the Summary Pretask and the +1 Demo as a whole because the motivation for the Summary Pretask was to shorten the length of the prompt.

\subsection{Summary Pretask} 
\label{section:ablation_summary_pretask}

% In order to analyze the effectiveness of each module in \model{}, we compare the \model{} with its variants under different settings. 

For the summary pretask, we apply two strategies: \textbf{All} and \textbf{Demo Only}.
The {Demo Only} strategy involves summarizing only the demonstrations used in the prompt, as described in Section~\ref{section:prompt_construction}. 
The All strategy builds on Demo Only by additionally summarizing descriptions of the target schema pairs.

As shown in Table \ref{table:ablation_summary_pretask}, the compression strategy of Demo Only outperforms the All strategy by 1.3\% on F1-socre under the 4-shot setting. This indicates that focusing compression on demonstration examples is more effective than extending it to target schema pairs.

\begin{table}[!htbp]
\centering
\footnotesize
\begin{tabular}{r|rr}
\toprule
Strategy	 	&	Accuracy	&	F1-score \\
\midrule
All   & {0.9892} & {0.2727} \\
Demo only & \textbf{0.9899} & \textbf{0.2857} \\
\bottomrule
\end{tabular}
\caption{Ablation study of the summary pretask's strategies on \textsf{Synthea}.}
\label{table:ablation_summary_pretask}
\end{table}

\subsection{Examples as Knowledge (EaK)} 

Following the \textbf{Demo Only} setting in Appendix~\ref{section:ablation_summary_pretask}, we further verify the validity of the knowledge constructed through EaK.
We introduce EaK as a knowledge source and form three distinct knowledge sets: \textsc{Wikidata}, EaK, and \textsc{Wikidata}+EaK, which are used for constructing the prompt. 

As shown in Table~\ref{table:ablation_eak}, EaK, characterized by its concise and high-quality knowledge, achieves the best performance among the three configurations. Using EaK alone as the knowledge source outperforms \textsc{Wikidata} and \textsc{Wikidata}+EaK by 0.7\% and 2.6\%, respectively, in terms of F1-score. This highlights the advantage of leveraging high-quality, compact examples as a knowledge source.
% As shown in Table~\ref{table:ablation_eak}, due to the short length and high quality of the EaK knowledge, individual EaK as the knowledge set achieves the best results, outperforming single \textsc{Wikidata} and \textsc{Wikidata}+EaK 0.7\% and 2.6\% on F1-score.

\begin{table}[!htbp]
\centering
\footnotesize
\begin{tabular}{r|rr}
\toprule
Knowledge Source	 	&	Accuracy	&	F1-score \\
\midrule
\textsc{Wikidata} 				& \underline{0.9899} & \underline{0.2857} 		\\ 
EaK 			   & \textbf{0.9902} & \textbf{0.2927}		\\
\textsc{Wikidata}+EaK 			 & 0.9889 & 0.2667 		\\
\bottomrule
\end{tabular}
\caption{Ablation study of EaK on \textsf{Synthea}.}
\label{table:ablation_eak}
\end{table}

\subsection{Self-indicator Reasoning (SIR)} 
\label{section:ablation_self_indicator_reasoning}

To evaluate the effectiveness and generalizability of Self-Indicator Reasoning (SIR), we introduce it into demonstrations across different knowledge sources for both the \textsf{MMM} and \textsf{Synthea} datasets.

\paragraph{Results on \textsf{MMM}}
Table~\ref{table:ablation_sir_MMM} presents the results with and without SIR on the \textsf{MMM} dataset with \textsc{Wikidata}, \textsc{Wikipedia}, and EaK as the knowledge sources, respectively. 
We observe that all cases with SIR outperform cases without SIR, with their F1-scores improving by an average of 7.73\%. More specifically, F1-scores of SIR with \textsc{Wikidata}, \textsc{Wikipedia}, and EaK as knowledge sources improved by 8.43\%, 7.05\%, and 7.70\%, respectively. These results demonstrate the robustness and effectiveness of SIR in enhancing the reasoning capabilities of the model.

\begin{table}[!htbp]
\centering
\footnotesize
\begin{tabular}{rr|rr}
\toprule
Knowledge Source 	& w. SIR &	Accuracy  &	F1-score \\
\midrule
\textsc{Wikidata}&   No   & 0.9498 & 0.6534 \\ 
\textsc{Wikidata}&   Yes   & 0.9667 & 0.7387 \\
\textsc{Wikipedia}   &   No   & 0.9603 & 0.6979 \\
\textsc{Wikipedia}   &   Yes   & \underline{0.9719} & \underline{0.7684} \\ 
EaK &   No   & 0.9593 & 0.6997 \\
EaK &   Yes   & \textbf{0.9729} & \textbf{0.7767} \\
\bottomrule
\end{tabular}
\caption{The ablation study of SIR on \textsf{MMM} (2-shot setting).}
\label{table:ablation_sir_MMM}
\end{table}

\paragraph{Results on \textsf{Synthea}}
Meanwhile, for \textsf{Synthea}, also based on the \emph{Demo Only} setting, we introduce SIR with \textsc{Wikidata} as the knowledge source. As shown in Table~\ref{table:ablation_sir_synthea}, SIR improved the F1-score on \textsf{Synthea} by 2.68\%, further confirming its effectiveness.

\begin{table}[!htbp]
\centering
\footnotesize
\begin{tabular}{rr|rr}
\toprule
Knowledge Source & w. SI 			&	Accuracy	&	F1-score \\
\midrule
\textsc{Wikidata}		& No	& {0.9899} 	& {0.2857}			\\
\textsc{Wikidata}		& Yes	&	\textbf{0.9926}	&	\textbf{0.3125}			\\
\bottomrule
\end{tabular}
\caption{The ablation study of SIR on \textsf{Synthea} (4-shot setting).}
\label{table:ablation_sir_synthea}
\end{table}

\subsection{Inconsistency-tolerant Generation Ensemble (IntGE)} 

The ablation experiments for IntGE build upon the findings from Appendix~\ref{section:ablation_self_indicator_reasoning}. Here, we evaluate the performance of IntGE by combining different knowledge sources and testing various configurations. Given that the impact of SIR is less pronounced on \textsf{Synthea} compared to \textsf{MMM}, we also include SIR as an experimental variable.

\paragraph{Results on \textsf{Synthea}}
Table~\ref{table:ablation_intge_synthea} showcases the results of IntGE applied to \textsf{Synthea} using different combinations of knowledge sources. All ensembled configurations outperform the single-knowledge-source baseline (\textsc{Wikidata}). 
The F1-scores of IntGE under combinations [NULL, \textsc{Wikipedia}+EaK, EaK], [NULL, \textsc{Wikipedia}, EaK], [\textsc{Wikidata}, \textsc{Wikipedia}+EaK, EaK], and [\textsc{Wikidata}, \textsc{Wikipedia}, EaK] are improved by 7.74\%, 4.04\%, 11.04\%, 0.33\%, respectively.
These results highlight the effectiveness of combining multiple knowledge sources in improving model performance.

\begin{table}[!htbp]
\centering
\footnotesize
\begin{adjustbox}{max width=\columnwidth}
\begin{tabular}{rr|rr}
\toprule
Knowledge Source & Voting & Accuracy & F1-score \\
\midrule
\textsc{Wikidata} & No & \underline{0.9926} & 0.3125 \\
{[NULL, \textsc{Wikipedia}+EaK, EaK]} & Yes & \underline{0.9926} & \underline{0.3889} \\
{[NULL, \textsc{Wikipedia}, EaK]} & Yes & \underline{0.9926} & 0.3529 \\
{[\textsc{Wikidata}, \textsc{Wikipedia}+EaK, EaK]} & Yes & \textbf{0.9936} & \textbf{0.4242} \\
{[\textsc{Wikidata}, \textsc{Wikipedia}, EaK]} & Yes & 0.9912 & 0.3158 \\
\bottomrule
\end{tabular}
\end{adjustbox}
\caption{Ablation study of IntGE on \textsf{Synthea} (4-shot setting). NULL denotes no knowledge source.}
\label{table:ablation_intge_synthea}
\end{table}

\paragraph{Results on \textsf{MMM}}
We conducted similar experiments on the \textsf{MMM} dataset, as shown in Table \ref{table:ablation_intge_MMM}. The F1-scores of IntGEs based on the combinations [\textsc{Wikidata}, \textsc{Wikipedia}, EaK], [\textsc{Wikidata}, \textsc{Wikipedia}, EaK$^*$], [\textsc{Wikidata}$^*$, \textsc{Wikipedia}, EaK], [\textsc{Wikipedia}, EaK, NULL] increased by 3.61\%, 2.88\%, 3.71\%, and 3.71\%, respectively, as compared to the results based on the single \textsc{Wikidata} data source.
Here, the asterisk ($^*$) indicates that the prompt for the corresponding knowledge source excludes self-indicators in reasoning steps. These results confirm the robustness and adaptability of IntGE across varying configurations and datasets.

\begin{table}[!htbp]
\centering
\footnotesize
\begin{adjustbox}{max width=\columnwidth}
\begin{tabular}{rr|rr}
\toprule
Knowledge Source  &   Voting   &	Accuracy &	F1-score \\
\midrule
\textsc{Wikidata}			                &   No  & 0.9667  & 0.7387			 \\
{[\textsc{Wikidata}, \textsc{Wikipedia}, EaK]}        &   Yes  & \textbf{0.9727} & \underline{0.7748}  \\ 
{[\textsc{Wikidata}, \textsc{Wikipedia}, EaK$^*$]}    &   Yes  & \underline{0.9715} & 0.7675  \\
{[\textsc{Wikidata}$^*$, \textsc{Wikipedia}, EaK]}    &   Yes  & \textbf{0.9727} & \textbf{0.7758}			 \\
{[\textsc{Wikipedia}, EaK, NULL]}            &   Yes  & \textbf{0.9727} & \textbf{0.7758}  	 \\
\bottomrule
\end{tabular}   
\end{adjustbox}
\caption{Ablation study of IntGE on \textsf{MMM} (2-shot setting).}
\label{table:ablation_intge_MMM}
\end{table}

\paragraph{Summary}
\added{The experiments here demonstrate that IntGE consistently improves performance by leveraging diverse knowledge sources, with significant F1-score gains across both \textsf{MMM} and \textsf{Synthea}. The results emphasize the value of combining high-quality knowledge sources and adapting reasoning strategies for optimal results.}

\section{Cost Analysis}
\label{appendix-section:cost_analysis}

\added{This section presents an ablation study of the main components of \model{}, analyzing their token costs and contributions to performance improvement. The results are grouped by techniques to evaluate their trade-offs in terms of cost and gain, providing deeper insights into the effectiveness of each component.}

\begin{figure*}[!htbp]
    \centerline{\includegraphics[width=2.25\columnwidth]{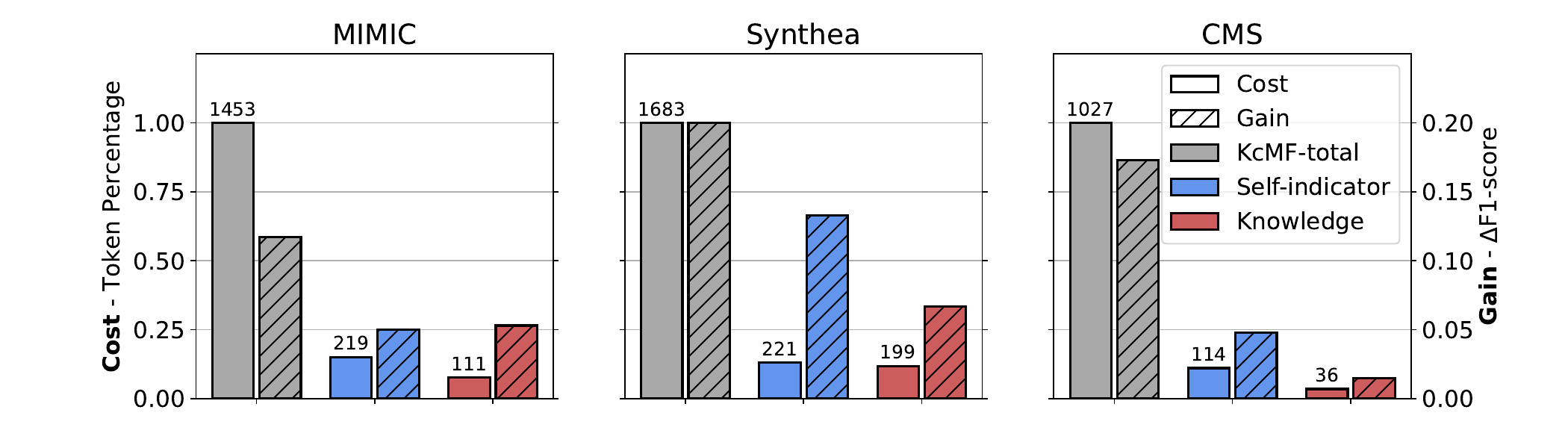}}
    \caption{\textcolor{newtabcap}{Cost and gain of the \model's main components, using \texttt{GPT-4o-mini} as the backbone model.}}
    \label{figure:cost_ablation}
\end{figure*}

\added{In Figure~\ref{figure:cost_ablation}, token costs and performance gain of \texttt{GPT-4o-mini} versioned \model{} are visualized as multiple bars, those unhatched represent the percentage of tokens the corresponding technique takes, while hatched bars represent the gain in F1-score of each technique to the baseline. 
It should be noted that the performance gain here and in the following text are obtained by ablations, e.g., the gain of the self-indicator is the difference between versions of \model{} with and without self-indicators. 
According to Figure~\ref{figure:cost_ablation}, both self-indicator and knowledge show consistent and independent performance gain, among which, we observe that self-indicator is significantly cost-effective, especially for schema matching tasks, which may be due to its effect in clarifying the complexity and heterogeneity of schemata. Also, knowledge is necessary for \model{} to achieve its optimal performance.}

\begin{table*}[ht]
\centering
\footnotesize
\begin{tabular}{cccccccc}
\toprule
\rowcolor{newtab!10}
            & \textsf{MIMIC}           & \textsf{Synthea}             & \textsf{CMS}               & \textsf{MMM}               & \textsf{AG}                & \textsf{WA}                & \textsf{DS}      \\
\midrule
\rowcolor{newtab!10}
Baseline    & 0.0077\textcent   & 0.0068\textcent   & 0.0074\textcent   & 0.0048\textcent   & 0.0034\textcent   & 0.0054\textcent   & 0.0056\textcent   \\
\rowcolor{newtab!10}
\model{}    & 0.0260\textcent   & 0.0278\textcent   & 0.0253\textcent   & 0.0207\textcent   & 0.0207\textcent   & 0.0271\textcent   & 0.0220\textcent   \\
\bottomrule
\end{tabular}   
\caption{\textcolor{newtabcap}{API costs using \texttt{GPT-4o-mini} endpoint.}}
\label{table:api_cost}
\end{table*}

\added{The token cost and gain of the F1-score of each technique are listed in Table~\ref{table:cost_ablation}, and the API invoking costs are reported in Table~\ref{table:api_cost}. Generally, the average cost of \model{} is times the baseline, due to the pseudo-code, reasoning steps, and the retrieved knowledge included in its query, and we consider this overhead worthwhile because 1) \model's total cost is still satisfactory, where around 60 cents for each dataset, and 2) it shows consistent and competitive performance gain towards different LLMs.}

\added{Further, we introduce the cost performance~(CP), which is the ratio of the gain of F1-score to the token percentage~(of total) for the main components of the \model{}, to measure the trade-off between cost and gain for techniques in \model{}. CP is formally defined in Equation~\ref{equation:cp}, where $|\mathcal{T'}|$ and $|\mathcal{T}|$ denote numbers of tokens corresponding to the single technique and the whole prompt respectively.}

\begin{equation}
\label{equation:cp}
    CP = \frac{\Delta F1}{{|\mathcal{T'}|}/{|\mathcal{T}|}}
\end{equation}

\added{To ensure fairness, CPs are averages calculated on \textsf{Synthea}, \textsf{MIMIC}, \textsf{CMS}, and \textsf{MMM} due to the absence of knowledge in the other three datasets. According to Table~\ref{table:cost_ablation}, we find that self-indicator is more cost-effective than knowledge, but knowledge is also necessary for the whole \model{} to achieve its best performance. Also, we observe that the self-indicator generalizes well on \textsf{AG}, \textsf{WA}, and \textsf{DS}, which indicates that self-indicator is a versatile method for improving LLM's performance on matching tasks.}

\begin{table}[!htbp]
\centering
\footnotesize
\begin{tabular}{cc|cccc}
\toprule
\rowcolor{newtab!10}
                                    &           & \multicolumn{4}{c}{Token Cost/Query} \\
\rowcolor{newtab!10}
\multicolumn{2}{c|}{\multirow{-2}{*}{Dataset}}  & K        & SI            & Infer   & Total \\
\midrule
\rowcolor{newtab!10}
                                    &   Cost   & 111      & 219           & 1123     & 1453     \\
\rowcolor{newtab!10}                         
\multirow{-2}{*}{\textsf{MIMIC}}    &   Gain   & $5.3\%$  & $5.0\%$          & -       & $11.7\%$  \\
\midrule
\rowcolor{newtab!10}
                                    &   Cost   & 199      & 221           & 1263     & 1683     \\
\rowcolor{newtab!10}
 \multirow{-2}{*}{\textsf{Synthea}}     &   Gain   & $6.7\%$  & $13.3\%$       & -       & $20.0\%$ \\
\midrule
\rowcolor{newtab!10}
 &   Cost   & 36       & 114           & 877     & 1027      \\
\rowcolor{newtab!10}                         
\multirow{-2}{*}{\textsf{CMS}}     &   Gain   & $1.5\%$  & $4.8\%$       & -       & $17.3\%$  \\
\midrule
\rowcolor{newtab!10}
    &   Cost   & 23       & 61            & 1051    & 1135      \\
\rowcolor{newtab!10}                         
\multirow{-2}{*}{\textsf{MMM}}   &   Gain   & $0.2\%$  & $0.3\%$          & -       & $8.2\%$   \\
\midrule
\rowcolor{newtab!10}
 &   Cost   & \multirow{2}{*}{-}       & 90            & 795     & 885     \\
\rowcolor{newtab!10}                         
\multirow{-2}{*}{\textsf{AG}}   &   Gain   &          & $1.2\%$       & -       & $15.6\%$     \\
\midrule
\rowcolor{newtab!10}
 &   Cost   & \multirow{2}{*}{-}       & 109           & 1044    & 1153    \\
\rowcolor{newtab!10}                         
\multirow{-2}{*}{\textsf{WA}}  &   Gain   &          & $2.2\%$       & -       & $41.0\%$     \\
\midrule
\rowcolor{newtab!10}
 &   Cost   & \multirow{2}{*}{-}       & 94            & 912     & 1006    \\
\rowcolor{newtab!10}                         
\multirow{-2}{*}{\textsf{DS}}    &   Gain   &          & $0.2\%$       & -       & $19.9\%$     \\
\midrule
\rowcolor{newtab!10}
&   CP                       & 0.441    & 0.474    & \multicolumn{2}{c}{-}        \\
\bottomrule
\end{tabular}   
\caption{\textcolor{newtabcap}{Cost and gain ablation results on all datasets in the main experiment, with averaged cost performance~(CP) for knowledge~(K) and self-indicator~(SI).}}
\label{table:cost_ablation}
\end{table}

% \balance
\onecolumn

\section{Discrepancies in LLM's Understanding of Schema Matching Instructions}
\label{appendix-section:gpt_response_sm}

\added{Table~\ref{appendix-table:ambiguity} presents the prompt using vanilla zero-shot Chain-of-Thought on \textsf{Synthea} and the response from \texttt{GPT-3.5}.
It can be observed that there is a discrepancy between
what \texttt{GPT-3.5} understands the instruction ``\texttt{match}'' and the actual objective of the schema matching task.}

\begin{table}[!htbp]
\centering
\footnotesize
\begin{tabular}{|p{\columnwidth}|}
\hline

Are schema A and B matched? Let's think step by step. \\
\textbf{schema A}: person-month\textunderscore of\textunderscore birth \\
\textbf{description of schema A}: the person domain contains records that uniquely identify each patient in the source data who is time at-risk to have clinical observations recorded within the source systems.;the month of birth of the person. for data sources that provide the precise date of birth, the month is extracted and stored in this field. \\
\textbf{schema B}: patients-birthdate \\
\textbf{description of schema B}: patient demographic data.;the date the patient was born. \\
\midrule
Based on the descriptions provided: \\ \\

**Schema A:** \\
- **Name:** person-month\textunderscore of\textunderscore birth \\
- **Description:** This schema appears to be related to a person domain where records uniquely identify each patient. It specifically captures the month of birth of the person. If the source data provides the precise date of birth, only the month is extracted and stored in this field. \\ \\

**Schema B:** \\
- **Name:** patients-birthdate \\
- **Description:** This schema relates to patient demographic data and specifically captures the date when the patient was born. \\ \\ 

Now, to determine if schema A and schema B are a match, we need to consider the following points: \\ \\

1. **Data Captured:** \\
   - Schema A captures only the month of birth if the precise date is available. \\
   - Schema B captures the full birthdate. \\ \\

2. **Granularity:** \\
   - Schema A has a lower granularity by capturing only the month, whereas Schema B captures the full date. \\

3. **Context and Usage:** \\
   - Schema A seems more focused on situations where the precise date is not necessary, and only the month suffices (possibly for anonymization or simplification purposes). \\
   - Schema B is more comprehensive as it captures the exact birthdate, which is typically more useful for demographic analysis and age calculation. \\ \\

**Conclusion:** \\
Based on the descriptions: \\
- Schema A (`person-month\textunderscore of\textunderscore birth`) and Schema B (`patients-birthdate`) do not appear to be a direct match because they capture different levels of detail regarding the birthdate of patients. Schema A focuses on the month of birth, while Schema B captures the full birthdate. \\ \\

Therefore, they serve different purposes and would likely be used in different contexts within a database or system that manages patient or person-related data. \\
\hline
\end{tabular}
\caption{\texttt{GPT-3.5}'s response on \textsf{Synthea}.}
\label{appendix-table:ambiguity}
\end{table}

\section{Detailed Prompts}
\label{appendix-section:prompts}

\added{In this section, we present all prompt templates used in the experiments covered in the main text, specifically:}

\begin{itemize}
    \item {\textbf{\Cref{appendix-table:keywords_extraction,appendix-table:keywords_quality_sm,appendix-table:keywords_quality_em,appendix-table:si_extraction,appendix-table:dos,appendix-table:retrieval_summarization}}}
    \begin{itemize}
        \item{Prompts for pre-tasks, including keywords extraction and quality management~(DaK introduced in Section~\ref{section:domain_knowledge_set}), self-indicator extraction, demonstration and retrieval summarization~(Section~\ref{section:prompt_construction}).}
    \end{itemize}
    \item {\textbf{\Cref{appendix-table:baseline_sm,appendix-table:baseline_em}}}
    \begin{itemize}
        \item{Prompts of baselines for comparison in the main experiment, \textbf{\Cref{appendix-table:baseline_sm,appendix-table:baseline_em}} are for SM and EM, separately.}
    \end{itemize}
    \item {\textbf{\Cref{appendix-table:pseudo_code_sm,appendix-table:pseudo_code_em}}}
    \begin{itemize}
        \item{Task-specific pseudo-codes for SM and EM used by \model{}.}
    \end{itemize}
    \item {\textbf{\Cref{appendix-table:kcmf_sm,appendix-table:kcmf_em}}}
    \begin{itemize}
        \item{Prompts obtained from Section~\ref{section:prompt_construction}, which are used to query LLMs directly.}
    \end{itemize}
\end{itemize}

\begin{table}[!htbp]
\centering
\footnotesize
\begin{tabular}{|p{\columnwidth}|}
\hline
\textbf{You need to extract all the keywords in the schema that require special domain knowledge to understand, keywords should be separated by commas.} \\

\textbf{Example 1}: \\
\textbf{Schema}: provider-npi \\
\textbf{Schema description}: the provider table contains a list of uniquely identified healthcare providers. these are individuals providing hands-on healthcare to patients, such as physicians, nurses, midwives, physical therapists etc.;the national provider identifier (npi) of the provider. \\
\textbf{Answer}: national provider identifier, npi \\ \\

\textbf{Example 2}: \\
\textbf{Schema}: imaging\textunderscore studies-sop description \\
\textbf{Schema description}: patient imaging metadata.;description of the sop code. \\
\textbf{Answer}: sop \\ \\

\textbf{Example 3}: \\
\textbf{Schema}: procedure\textunderscore occurrence-modifier\textunderscore concept\textunderscore id \\
\textbf{Schema description}: the procedure\textunderscore occurrence table contains records of activities or processes ordered by, or carried out by, a healthcare provider on the patient to have a diagnostic or therapeutic purpose. procedures are present in various data sources in different forms with varying levels of standardization.;a foreign key to a standard concept identifier for a modifier to the procedure (e.g. bilateral). these concepts are typically distinguished by 'modifier' concept classes (e.g., 'cpt4 modifier' as part of the 'cpt4' vocabulary). \\
\textbf{Answer}: foreign key, identifier, cpt4 \\ \\

\textbf{Your turn}: \\
\hline
\end{tabular}
\caption{Prompt for keyword extraction.}
\label{appendix-table:keywords_extraction}
\end{table}

\begin{table}[!htbp]
\centering
\footnotesize
\begin{tabular}{|p{\columnwidth}|}
\hline
\textbf{You need to find out which of the given keywords are relevant to the database or medical field and return them, keywords should be separated by commas.} \\ \\

\textbf{Example 1}: \\
birthday, home, dcm, location, primary key \\
\textbf{Answer}: dcm, primary key \\ \\

\textbf{Example 2}: \\
endtime, id, recognition, observation, data model, algorithm, artificial \\
\textbf{Answer}: id, data model \\ \\

\textbf{Your turn}: \\
\hline
\end{tabular}
\caption{Prompt for keyword quality management for the schema matching datasets (\textsf{MIMIC}, \textsf{Synthea}, and \textsf{CMS}).}
\label{appendix-table:keywords_quality_sm}
\end{table}

\begin{table}[!htbp]
\centering
\footnotesize
\begin{tabular}{|p{\columnwidth}|}
\hline
\textbf{Tell if the given word is a hard-to-understand biomedical domain term, only yes or no.} \\ \\

\textbf{Example 1}: \\
leaf \\
\textbf{Answer}: no \\ \\

\textbf{Example 2}: \\
deoxynivalenol \\
\textbf{Answer}: yes \\ \\

\textbf{Your turn}: \\
\hline
\end{tabular}
\caption{Prompt for keywords quality management for the entity matching datasets (\textsf{MMM}, \textsf{AG}, \textsf{WA}, and \textsf{DS}).}
\label{appendix-table:keywords_quality_em}
\end{table}

\begin{table}[!htbp]
\centering
\footnotesize
\begin{tabular}{|p{\columnwidth}|}
\hline
\textbf{Instruction}: Given two schemas, you need to summarize the column and table for each considering the given knowledge.    \\
{[\textit{Examples}]} \\
\textbf{Your turn}:      \\
$\{r,r'\}$      \\
$[K_i]$        \\
\textbf{Answer}:         \\
\hline
\end{tabular}
\caption{Prompt for self-indicator extraction (SIR).}
\label{appendix-table:si_extraction}
\end{table}

\begin{table}[!htbp]
\centering
\footnotesize
\begin{tabular}{|p{\columnwidth}|}
\hline
\textbf{Instruction}: You need to summarize the given schema based on its schema name and description. The summary should be focused on retaining and explaining concepts in the database domain. Schema name is the table and column names of the schema separated by a dash. Schema description is the table and column descriptions of the schema separated by a semicolon. \\\\

{[\textit{Examples}]}       \\\\

\textbf{Your turn}:         \\
$\{r,r'\}$                  \\
\textbf{Answer}:            \\
\hline
\end{tabular}
\caption{Prompt for demonstration summarization.}
\label{appendix-table:dos}
\end{table}

\begin{table}[!htbp]
\centering
\footnotesize
\begin{tabular}{|p{\columnwidth}|}
\hline
\textbf{Instruction}: You need to summarize the given text into a paragraph less than 1000 words. \\
\{CONTENT\}             \\
\textbf{Your answer}:   \\
\hline
\end{tabular}
\caption{Prompt for retrieval summarization.}
\label{appendix-table:retrieval_summarization}
\end{table}

\begin{table}[!htbp]
\centering
\footnotesize
\begin{tabular}{|p{\columnwidth}|}
\hline
\textbf{Are schema A and B the same? ONLY yes or no.} \\
{[\textit{Examples}]} \\
\textbf{Your turn}: \\
$\{r, r'\}$ \\
\{\textit{RESPONSE}\} \\
\hline
\end{tabular}
\caption{Prompt template of baseline for the schema matching datasets (\textsf{MIMIC}, \textsf{Synthea}, and \textsf{CMS}).}
\label{appendix-table:baseline_sm}
\end{table}

\begin{table}[!htbp]
\centering
\footnotesize
\begin{tabular}{|p{\columnwidth}|}
\hline
\textbf{Question: Do entity A and entity B refer to the same real-world concept? Only yes or no.} \\
{[\textit{Examples}]} \\
\textbf{Your turn}:  \\
$\{r,r'\}$ \\
\textbf{Answer}: \\
\{\textit{RESPONSE}\} \\
\hline
\end{tabular}
\caption{Prompt template of baseline for the entity matching datasets (\textsf{MMM}, \textsf{AG}, \textsf{WA}, and \textsf{DS}).}
\label{appendix-table:baseline_em}
\end{table}

 \begin{table}[!htbp]
\centering
\footnotesize
\begin{tabular}{|p{\columnwidth}|}
\hline
I: Rules II, III, and IV MUST be checked SEQUENTIALLY until you conclude an answer. \\
II: If the columns of the two schemas can not be the same type of data in the database, the answer is no, otherwise, check rule III. \\
III: If the tables of the two schemas are not semantically the same, the answer is no, otherwise, check rule IV. \\
IV: If the columns of the two schemas do not refer to the same concept, the answer is no, otherwise, the answer is yes. \\
\hline
\end{tabular}
\caption{Pseudo-code for the schema matching datasets (\textsf{MIMIC}, \textsf{Synthea}, and \textsf{CMS}).}
\label{appendix-table:pseudo_code_sm}
\end{table}

\begin{table}[!htbp]
\centering
\footnotesize
\begin{tabular}{|p{\columnwidth}|}
\hline
I: Rules II, III, and IV MUST be checked SEQUENTIALLY until you conclude an answer. \\
II: If Entity A is an abbreviation of Entity B or vice versa, the answer is yes, otherwise check rule III. \\
III: If Entity A is an alias of Entity B or vice versa, the answer is yes, otherwise check rule IV. \\
IV: If Entity A and Entity B refer to the same real-world concept the answer is yes, otherwise the answer is no. \\
\hline
\end{tabular}
\caption{Pseudo-code for the entity matching datasets (\textsf{MMM}, \textsf{AG}, \textsf{WA}, and \textsf{DS}).}
\label{appendix-table:pseudo_code_em}
\end{table}

\begin{table}[htbp]
\centering
\footnotesize
\begin{tabular}{|p{\columnwidth}|}
\hline
\textbf{Question}: \\
Can records in schema B be transformed and stored in schema A? The task should be solved by completing the reasoning steps and concluding a final answer ONLY yes or no. Do not stop until you draw a final answer. Schema name is the table and column names of the schema separated by a dash. \\

\textbf{Rules for the task}: \\
{[$\mathcal{R}$]} \\

{[\textit{Demonstrations}]} \\

\textbf{Your turn}: \\
$\{r, r'\}$ \\
\textbf{Knowledge for the task}: \\
{[$K_i$]} \\
\textbf{Reasoning}: \\
1. \{\textit{Self-indicator}\} \\
\textbf{Please continue the reasoning until you draw a final answer ONLY yes or no}: \\
\{\textit{RESPONSE}\} \\
\hline
\end{tabular}
\caption{Prompt template of \model{} for the schema matching datasets (\textsf{MIMIC}, \textsf{Synthea}, and \textsf{CMS}).}
\label{appendix-table:kcmf_sm}
\end{table}

\begin{table}[htbp]
\centering
\footnotesize
\begin{tabular}{|p{\columnwidth}|}
\hline
\textbf{Question}: \\
Do entity A and entity B refer to the same real-world concept? You must think step by step, and finally draw an answer only yes or no. \\

\textbf{Rules for the task}: \\
{[$\mathcal{R}$]} \\

{[\textit{Demonstrations}]} \\

\textbf{Your turn}: \\
$\{r, r'\}$ \\
\textbf{Knowledge for the task}: \\
{[$K_i$]} \\
\textbf{Reasoning}: \\
1. \{\textit{Self-indicator}\} \\
\textbf{Please continue the reasoning until you draw a final answer ONLY yes or no}: \\
\{\textit{RESPONSE}\} \\
\hline
\end{tabular}
\caption{Prompt template of \model{} for the entity matching datasets (\textsf{MMM}, \textsf{AG}, \textsf{WA}, and \textsf{DS}).}
\label{appendix-table:kcmf_em}
\end{table}

\end{document}